\documentclass[a4paper,10pt,oneside,fleqn,preprint]{elsarticle}
\usepackage[%
paperwidth=192mm,
paperheight=262mm,
vmargin={19mm,19mm},
hmargin={13.7mm,13.7mm},
headsep=12pt,
footskip=12pt,
]{geometry}

\usepackage{lmodern}
\usepackage{eulervm}
\usepackage{fourier}
\usepackage[caption=false,font=footnotesize,labelfont=sf,textfont=sf]{subfig}
\usepackage{algorithmic}
\usepackage[utf8]{inputenc}
\usepackage[T1]{fontenc}
\usepackage[usenames,dvipsnames]{xcolor}
\usepackage{graphicx}
\usepackage{textcomp}
\usepackage{standalone}
\usepackage[final]{pdfpages}
\usepackage{calc}
\usepackage[normalem]{ulem}
\usepackage{tikz,pgfplots}
\usepackage{grffile}
\usepackage{amsmath,amssymb,amsfonts}
\usepackage{mathtools}
\usepackage{esdiff}
\usepackage{commath}

\usepackage{breqn}
\usepackage{gensymb}
\usepackage{url}

\usepackage{tabularx}
\usepackage{array}
\usepackage{xspace}
\usepackage{verbatim}
\usepackage{booktabs}
\usepackage{blindtext}
\usepackage[flushleft]{threeparttable}
\usepackage[shortcuts]{extdash}
\usepackage{fancyref}
\usepackage{colortbl}
\usepackage{placeins}
\usepackage{siunitx}
\DeclareSIUnit{\pii}{\ensuremath{\pi}}
\newcommand*{\ThresholdLow}{0.001}
\newcommand*{\ThresholdHigh}{100000}
\usepackage{expl3}
\ExplSyntaxOn
\cs_new_eq:NN \fpcmpTF \fp_compare:nTF
\ExplSyntaxOff
\let\OldNum\num%
\renewcommand*{\num}[2][]{%
    \fpcmpTF{abs(#2)<=\ThresholdLow}{%
        \OldNum[round-mode=figures,round-precision=2,scientific-notation=true,#1]{#2}%
    }{%
        \fpcmpTF{abs(#2)>=\ThresholdHigh}{%
            \OldNum[round-mode=figures,round-precision=2,scientific-notation=true,#1]{#2}%
        }{%
            \OldNum[scientific-notation=false,#1]{#2}%
        }%
    }%
}%

\newcommand{\etal}{\textit{et al}.\xspace}

\newcommand{\eg}{\textit{e}.\textit{g}.,\xspace}
\usepackage{array}
\newcolumntype{L}[1]{>{\raggedright\let\newline\\\arraybackslash\hspace{0pt}}m{#1}}
\newcolumntype{C}[1]{>{\centering\let\newline\\\arraybackslash\hspace{0pt}}m{#1}}
\newcolumntype{R}[1]{>{\raggedleft\let\newline\\\arraybackslash\hspace{0pt}}m{#1}}

\newlength\figureheight
\newlength\figurewidth
\usepackage{mathtools}

\pgfplotsset{compat=newest}
\usetikzlibrary{shapes,arrows,calc,intersections,positioning,fit,plotmarks,arrows.meta}
\usepgfplotslibrary{patchplots}
\pgfplotsset{plot coordinates/math parser=false}

\usepackage[symbols]{glossaries-extra}
\glsdisablehyper
\setabbreviationstyle{long-short}
\setabbreviationstyle{long-short-desc}
\setglossarystyle{altlist}

\newabbreviation[description={robotic actuator whose effective mechanical stiffness can be changed dynamically}]{vsa}{VSA}{variable stiffness actuator}
\newabbreviation[description={robotic actuator whose effective mechanical impedance (stiffness and damping) can be changed dynamically}]{via}{VIA}{variable impedance actuator}
\newabbreviation[description={robotic actuator with an elastic element making the end effector compliant and that can store potential energy to be released as kinetic energy}]{sea}{SEA}{series elastic actuator}
\newabbreviation[description={measure for the severity of an impact on a human head}]{hic}{HIC}{head injury criterion}
\newabbreviation[description={neurological system based in the spinal cord and brain, amongst others responsible for low-level coordination of muscular stimulation}]{cns}{CNS}{central nervous system}

\glsxtrnewsymbol[description={resonance frequency}]{fresonance}{\ensuremath{f_0}\xspace}
\glsxtrnewsymbol[description={resonance angular velocity}]{wresonance}{\ensuremath{\omega_0}\xspace}
\glsxtrnewsymbol[description={imaginary unit such that $\imag^2 + 1 = 0$}]{imag}{\ensuremath{j}\xspace}
\glsxtrnewsymbol[description={stiffness of an elastic element}]{kstiffness}{\ensuremath{k}\xspace}
\glsxtrnewsymbol[description={mass of a body}]{mmass}{\ensuremath{m}\xspace}
\glsxtrnewsymbol[description={reference impedance}]{Znullvar}{\ensuremath{Z_0}\xspace}
\glsxtrnewsymbol[description={transfer function}]{Hvar}{\ensuremath{H}\xspace}
\glsxtrnewsymbol[description={}]{Kvar}{\ensuremath{K}\xspace}
\glsxtrnewsymbol[description={viscous damping coefficient}]{bdamping}{\ensuremath{b}\xspace}
\glsxtrnewsymbol[description={damping ratio}]{zetadamping}{\ensuremath{\zeta}\xspace}
\glsxtrnewsymbol[description={force}]{Fforce}{\ensuremath{F}\xspace}
\glsxtrnewsymbol[description={frequency}]{ffrequency}{\ensuremath{f}\xspace}
\glsxtrnewsymbol[description={damped resonance frequency}]{fdampedresonance}{\ensuremath{f_r}\xspace}
\glsxtrnewsymbol[description={end-effector angular velocity}]{thetadote}{\ensuremath{\dot{\theta}_{e}}\xspace}
\glsxtrnewsymbol[description={hammer angular velocity}]{thetadoth}{\ensuremath{\dot{\theta}_{h}}\xspace}
\glsxtrnewsymbol[description={peak end-effector angular velocity}]{thetadotemax}{\ensuremath{\dot{\theta}_{e,max}}\xspace}
\glsxtrnewsymbol[description={peak hammer angular velocity}]{thetadothmax}{\ensuremath{\dot{\theta}_{h,max}}\xspace}
\glsxtrnewsymbol[description={oscillation period}]{Tperiod}{\ensuremath{T}\xspace}
\glsxtrnewsymbol[description={gain}]{Ggain}{\ensuremath{G}\xspace}
\glsxtrnewsymbol[description={reference gain}]{Grefgain}{\ensuremath{G_{ref}}\xspace}
\glsxtrnewsymbol[description={relative gain}]{etarelgain}{\ensuremath{\eta}\xspace}
\glsxtrnewsymbol[description={frequency error}]{efreqerror}{\ensuremath{e}\xspace}
\glsxtrnewsymbol[description={hammer impedance}]{Zh}{\ensuremath{Z_h}\xspace}
\glsxtrnewsymbol[description={end-effector impedance}]{Ze}{\ensuremath{Z_e}\xspace}
\glsxtrnewsymbol[description={output impedance}]{Zto}{\ensuremath{Z_{to}}\xspace}
\glsxtrnewsymbol[description={force exerted by the human operator on the handle device as result of the arm impedance}]{fh}{\ensuremath{f_h}\xspace}
\glsxtrnewsymbol[description={force intentionally exerted by the human operator on the handle device}]{fhext}{\ensuremath{f_{\tilde{h}}}\xspace}
\glsxtrnewsymbol[description={force on the hammer as result of its impedance}]{fe}{\ensuremath{f_e}\xspace}
\glsxtrnewsymbol[description={external forces on the hammer}]{feext}{\ensuremath{f_{\tilde{e}}}\xspace}
\glsxtrnewsymbol[description={tool device force}]{fm}{\ensuremath{f_m}\xspace}
\glsxtrnewsymbol[description={force commanded by the controller to the handle device}]{fmc}{\ensuremath{f_{mc}}\xspace}
\glsxtrnewsymbol[description={force commanded by the controller to the tool device}]{fsc}{\ensuremath{f_{sc}}\xspace}
\glsxtrnewsymbol[description={handle device external force}]{fmext}{\ensuremath{f_{\tilde{m}}}\xspace}
\glsxtrnewsymbol[description={tool device force}]{fs}{\ensuremath{f_s}\xspace}
\glsxtrnewsymbol[description={time delay}]{Tdelay}{\ensuremath{T_d}\xspace}
\glsxtrnewsymbol[description={force}]{fst}{\ensuremath{f_{st}}\xspace}
\glsxtrnewsymbol[description={force}]{fsb}{\ensuremath{f_{sb}}\xspace}
\glsxtrnewsymbol[description={force}]{fsext}{\ensuremath{f_{\tilde{s}}}\xspace}
\glsxtrnewsymbol[description={force command}]{fCm}{\ensuremath{f_{C_m}}\xspace}
\glsxtrnewsymbol[description={force command}]{fCs}{\ensuremath{f_{C_s}}\xspace}
\glsxtrnewsymbol[description={force command}]{fCone}{\ensuremath{f_{C_1}}\xspace}
\glsxtrnewsymbol[description={force command}]{fCtwo}{\ensuremath{f_{C_2}}\xspace}
\glsxtrnewsymbol[description={}]{fCthree}{\ensuremath{f_{C_3}}\xspace}
\glsxtrnewsymbol[description={}]{fCfour}{\ensuremath{f_{C_4}}\xspace}
\glsxtrnewsymbol[description={}]{fCfive}{\ensuremath{f_{C_5}}\xspace}
\glsxtrnewsymbol[description={}]{fCsix}{\ensuremath{f_{C_6}}\xspace}
\glsxtrnewsymbol[description={}]{mh}{\ensuremath{m_h}\xspace}
\glsxtrnewsymbol[description={}]{mm}{\ensuremath{m_m}\xspace}
\newglossaryentry{tool device}
{
    name={tool device},
    description={remote device of the teleoperation system that interacts with the environment}
}
\newglossaryentry{handle device}
{
    name={handle device},
    description={local device of the teleoperation system with a haptic interface for the operator to interact with}
}
\newglossaryentry{accurate}
{
    name={accurate},
    first={accurate (high precision and trueness)},
    description={precise and with little difference to the true value}
}
\newglossaryentry{precise}
{
    name={precise},
    description={with little difference between measurements}
}
\newglossaryentry{stiffness}
{
    name={stiffness},
    description={metric for the force of a material resisting deformation calculated as force per deformation}
}
\newglossaryentry{virtual stiffness}
{
    name={virtual stiffness},
    description={stiffness that is rendered by a control loop through a mechanically stiff robot}
}
\newglossaryentry{effective stiffness}
{
    name={effective stiffness},
    description={stiffness at the end effector of a robot with mechanically variable stiffness that can be changed by changing the configuration of the stiffness variation mechanism}
}
\newglossaryentry{agonist and antagonist muscles}
{
    name={agonist and antagonist muscles},
    description={muscles pulling in opposing directions on a human articulation to move and position a limb in one direction or another}
}
\newglossaryentry{teleoperation}
{
    name={teleoperation},
    description={mode of operation where a human operator interacting with a handle device controls a robotic tool device in real-time, ideally able to manipulate objects in the operating environment as if he or she was interacting with the objects with his or her own hands}
}
\newglossaryentry{teleoperation system}
{
    name={teleoperation system},
    description={system for teleoperation, consisting of the handle device, the tool device and a communication link connecting the two}
}
\newglossaryentry{transparent}
{
    name={transparent},
    description={quality of a teleoperation system to make tool and handle device appear with equal force and position to their environment and the human operator at any time}
}
\newglossaryentry{compliant}
{
    name={compliant},
    description={quality of an actuator to have relatively low stiffness, typically in the order of magnitude that humans can achieve in their limbs, and well defined behavior under external loads}
}
\newglossaryentry{proprioceptive reflexes}
{
    name={proprioceptive reflexes},
    description={low-level control loop of the \gls{cns} to achieve certain motoric behavior}
}
\newglossaryentry{remote center of compliance device}
{
    name={remote center of compliance device},
    description={slightly elastic component placed at the tip of a robot end-effector to allow for small deformations in order to compensate for misalignments during placement tasks}
}

\newcommand{\nth}[1]{#1\ensuremath{^{th}}\xspace}
\newcommand{\percentile}[1]{\nth{#1} percentile}
\newcommand{\percentiles}[1]{\nth{#1} percentiles}
\newcommand{\siglevel}[1]{\SI{#1}{\percent}-level}

\newcommand{\legendRectangle}[1]{\begin{tikzpicture}\draw[fill=color#1] (0,0) rectangle (1em,1em); \end{tikzpicture}}

\usepackage{floatrow}
\usepackage[nooneline,footnotesize]{caption}

\definecolor{printer1}{cmyk}{0.10,0.90,0.80,0.0}%
\definecolor{printer2}{cmyk}{0.80,0.30,0.0,0.0}%
\definecolor{printer3}{cmyk}{0.70,0.0,0.80,0.0}%
\definecolor{printer4}{cmyk}{0.40,0.65,0.0,0.0}%
\definecolor{printer5}{cmyk}{0.0,0.50,1.00,0.0}%
\definecolor{printer6}{cmyk}{0.0,0.0,0.80,0.0}%
\definecolor{printer7}{cmyk}{0.35,0.60,0.80,0.0}%
\definecolor{printer8}{cmyk}{0.0,0.50,0.0,0.0}%
\definecolor{printer9}{cmyk}{0.0,0.0,0.0,0.40}%	
\definecolor{visible1}{cmyk}{0.35,0.07,0.0,0.0}%	
\definecolor{visible2}{cmyk}{0.90,0.30,0.0,0.0}%	
\definecolor{visible3}{cmyk}{0.30,0.0,0.45,0.0}%	
\definecolor{visible4}{cmyk}{0.80,0.0,1.00,0.0}%	
\definecolor{visible5}{cmyk}{0.0,0.40,0.25,0.0}%	
\definecolor{visible6}{cmyk}{0.10,0.90,0.80,0.0}%	
\definecolor{visible7}{cmyk}{0.0,0.25,0.50,0.0}%	
\definecolor{visible8}{cmyk}{0.0,0.50,1.00,0.0}%	
\definecolor{visible9}{cmyk}{0.20,0.25,0.0,0.0}%	
\definecolor{visible10}{cmyk}{0.60,0.70,0.0,0.0}%	
\definecolor{visible11}{cmyk}{0.0,0.0,0.40,0.0}%	
\definecolor{visible12}{cmyk}{0.23,0.73,0.98,0.12}%	
\definecolor{dark1}{cmyk}{0.90,0.0,0.55,0.0}%	
\definecolor{dark2}{cmyk}{0.15,0.60,1.00,0.0}%	
\definecolor{dark3}{cmyk}{0.55,0.45,0.0,0.0}%	
\definecolor{dark4}{cmyk}{0.05,0.85,0.05,0.0}%	
\definecolor{dark5}{cmyk}{0.60,0.10,1.00,0.0}%	
\definecolor{dark6}{cmyk}{0.10,0.30,1.00,0.0}%	
\definecolor{dark7}{cmyk}{0.35,0.45,0.90,0.0}%	
\definecolor{dark8}{cmyk}{0.0,0.0,0.0,0.60}%
\definecolor{gray1}{gray}{0.0}%	
\definecolor{gray2}{gray}{0.4}%	
\definecolor{gray3}{gray}{0.8}%	
\definecolor{matlab1}{rgb}{0.00000,0.44700,0.74100}%
\definecolor{matlab2}{rgb}{0.85000,0.32500,0.09800}%
\definecolor{matlab3}{rgb}{0.92900,0.69400,0.12500}%
\definecolor{matlab4}{rgb}{0.49400,0.18400,0.55600}%
\definecolor{matlab5}{rgb}{0.46600,0.67400,0.18800}%
\definecolor{matlab6}{rgb}{0.3,0.3,0.3}%
\definecolor{matlab7}{rgb}{0.30100,0.74500,0.93300}%
\definecolor{matlab8}{rgb}{0.63500,0.07800,0.18400}%
\definecolor{matlab9}{rgb}{0.7,0.7,0.7}%

\definecolor{targetbrushposition}{RGB}{200, 150, 30}%
\definecolor{targetbrushhammer}{RGB}{50, 200, 100}%
\definecolor{actualbrush}{RGB}{50, 150, 230}%
\definecolor{boxbrush}{RGB}{50, 200, 100}%

\colorlet{colorHighImp}{matlab1}
\colorlet{colorLowImp}{matlab2}
\colorlet{colorAdaptive}{matlab3}
\colorlet{colorTaskswitch}{matlab4}
\definecolor{colorHypConf}{cmyk}{0.3,0,0.3,0}
\definecolor{colorHypRej}{cmyk}{0,0.3,0.3,0}

\newlength{\dotSymRad}
\newlength{\hammerlength}
\newlength{\estoplength}
\newcommand{\junction}[2]{%
	\draw[fill] let \p1 = #1,\p2 = #2 in (\x1,\y2) circle (1pt);
}

\newlength{\gndsymwidth}
\newlength{\gndsymheight}
\newlength{\lindampersymheight}
\newlength{\lindampersymwidth}
\newlength{\linspringsymheight}
\newlength{\linspringsymwidth}
\newlength{\masssymrad}
\newcommand{\imag}{\ensuremath{j}\xspace}
\newcommand{\kstiffness}{\ensuremath{k}\xspace}

\newcommand{\mmass}{\ensuremath{m}\xspace}

\newcommand{\Zh}{\ensuremath{Z_h}\xspace}

\newcommand{\fh}{\ensuremath{f_h}\xspace}
\newcommand{\fhext}{\ensuremath{f_{\tilde{h}}}\xspace}
\newcommand{\fe}{\ensuremath{f_e}\xspace}
\newcommand{\feext}{\ensuremath{f_{\tilde{e}}}\xspace}

\newcommand{\fmc}{\ensuremath{f_{mc}}\xspace}

\newcommand{\vm}{\ensuremath{\dot{x}_m}\xspace}
\newcommand{\vmmax}{\ensuremath{\dot{x}_{m,max}}\xspace}
\newcommand{\vmmin}{\ensuremath{\dot{x}_{m,min}}\xspace}
\newcommand{\kmax}{\ensuremath{k_{max}}\xspace}
\newcommand{\kmin}{\ensuremath{k_{min}}\xspace}
\newcommand{\kfunc}[1]{\ensuremath{k\left(#1\right)}\xspace}
\newcommand{\Dfunc}[1]{\ensuremath{D\left(#1\right)}\xspace}
\newcommand{\bfunc}[1]{\ensuremath{b\left(#1\right)}\xspace}
\newcommand{\Dmax}{\ensuremath{D_{max}}\xspace}
\newcommand{\Dmin}{\ensuremath{D_{min}}\xspace}

\newcommand{\vs}{\ensuremath{\dot{x}_s}\xspace}

\newcommand{\slap}{\ensuremath{s}\xspace}

\newcommand{\Cm}{\ensuremath{C_m}\xspace}

\newcommand{\Cone}{\ensuremath{C_1}\xspace}
\newcommand{\Ctwo}{\ensuremath{C_2}\xspace}
\newcommand{\Cthree}{\ensuremath{C_3}\xspace}
\newcommand{\Cfour}{\ensuremath{C_4}\xspace}

\newcommand{\CZs}{\ensuremath{C_{Z_s}}\xspace}

\newcommand{\Zsset}{\ensuremath{z_{sc}}\xspace}
\newcommand{\Zs}{\ensuremath{Z_s}\xspace}

\newcommand{\TFmotor}{\ensuremath{TF_{motor}}\xspace}

\newcommand{\Zminv}{\ensuremath{Z_m^{-1}}\xspace}

\newcommand{\avar}{\ensuremath{a}\xspace}

\newcommand{\CI}[1][95]{\ensuremath{\text{CI}_{#1\%}}\xspace}

\newcommand{\ExpOne}{Experiment~1\xspace}

\journal{IEEE Transactions on Haptics}

\begin{document}
    
\begin{frontmatter}
\title{Self-Adapting Variable Impedance Actuator Control for Precision and Dynamic Tasks}
\author{Manuel Aiple\fnref{affbme}}
\author{André Schiele\fnref{affbme}}
\author{Frans C.T. van der Helm\fnref{affbme}}

\tnotetext[authoremail]{Corresponding author M. Aiple: \texttt{m.aiple@tudelft.nl}}
\address{Delft University of Technology, Mekelweg 2, 2628 CN Delft, Netherlands}
\fntext[affbme]{All authors are with the BioMechanical Engineering Department, Faculty of Mechanical, Maritime and Materials Engineering, Delft University of Technology, Netherlands.}

\begin{abstract}
   \Glsfirstplural{via} as tool devices for teleoperation could extend the range of tasks that humans can perform through a teleoperated robot by mimicking the change of upper limb stiffness that humans intuitively perform for different tasks and thereby increasing the dynamic range of the robot.
   However, this requires that the impedance be controlled appropriately, which is so far often achieved by sensors measuring the operator's muscle tension.
   But additional sensors mean increased system complexity and possibly longer donning time if the sensors need to be fixed and calibrated on the operator.
   Goal of this study is to show the effectiveness of a controller that does not require additional sensors, thereby reducing system complexity and increasing ease of use.
   The controller should allow to perform precise positioning tasks and dynamic tasks like hammering through teleoperation with a VIA tool device without requiring the human user to actively change the impedance setting of the VIA.
   This is achieved by a control law according to the principle ``slow-stiff/fast-soft'', tuned to the characteristics of human motion during precision tasks and dynamic tasks.
   The controller was tested in a human user study with 24 participants comparing the human-machine performance with the self-adapting controller in a bilateral telemanipulation experiment with two tasks (precision/dynamic) using three impedance settings (high/low/adaptive impedance).
   The results indicate that the proposed system performs equally well as state of the art stiff teleoperation devices for precision tasks, while having benefits in terms of increased safety and reduced wear for dynamic tasks.
   This is a step towards teleoperation with a wide dynamic range allowing to intuitively perform tasks in the full human motion bandwidth without requiring the human operator to select a specific mode of operation.
\end{abstract}
\begin{keyword}
    Human-Robot Interaction, Variable Impedance Actuator, Teleoperation
\end{keyword}

\end{frontmatter}

\section{Introduction}%
\label{sec:introduction}%

Variable Impedance Actuators (VIAs) \cite{Vanderborght2012,Vanderborght2013,Wolf2016} are a very interesting technology for teleoperation, as they allow to dynamically change the impedance of a robotic manipulator in a similar way to humans adapting the impedance of their arms to the task at hand \cite{Hogan1984}.
Compared to classical stiff robots, \glspl{via} thus have additional degrees of freedom (stiffness and damping) that need to be controlled on top of position control, which is not a trivial problem.
Specifically, the scope of this study was to develop a control law for the teleoperated dynamic robotic actuator (Dyrac) that we developed, optimized for highly dynamic tasks like throwing or hammering \cite{DyracPaper}.

For automatically executed tasks, optimal control laws for the impedance control have been presented, optimizing for maximum energy efficiency, maximum output velocity or other criteria \cite{Semini2015,Buchli2011,Braun2012}.
These are applicable to preplanned task execution.
For teleoperated robots however, an optimal control law cannot be determined the same way due to the unpredictable nature of the human trajectory that needs to be followed by the robot.
Thus, an impedance controller is required that works without knowing the full trajectory beforehand.

One approach that has been explored in previous research is to estimate the stiffness of the limbs of the human operator interacting with the handle device of the teleoperation system and reproduce it in the tool device.
This can be achieved for example through EMG measurement \cite{VanTeeffelen2018}, vibration based estimation \cite{Hill2009}, or grip force measurement \cite{Walker2010a}.
It assumes correct stiffness intention and good body control by the human operator and correct estimation of the operator's limbs' stiffness by the teleoperation system. It also requires additional sensors, some of them (\eg EMG) needing to be fixed and calibrated to the operator's body, thereby increasing the system complexity and preparation time before using the system, making it cumbersome to implement and use.

Goal of this paper is to experimentally validate a control law for \glspl{via} that does not require external impedance commands, but continuously adapts the tool actuator impedance according to the instantaneous motion command.
The control law is based on the ``slow-stiff/fast-soft'' principle inspired by ``Fast and "Soft-Arm" Tactics'' of Bicchi \etal \cite{Bicchi2004}.
It should allow efficient execution of precision tasks as well as highly dynamic tasks like hammering, throwing, shaking, jolting, etc., without requiring an explicit task selection.

We have shown in previous research through two experiments how \glspl{sea} could increase the output velocity compared to stiff actuators for highly dynamic tasks \cite{Aiple2018}.
But soft \glspl{sea} are not suitable for precision tasks as they are difficult to control in position and quickly start oscillating in free air.
However, precision tasks are crucial to teleoperation as they include very common tasks like pick-and-place tasks, insertion, inspection, etc.
In this paper we therefore show through a third experiment how precision tasks and highly dynamic tasks can be performed efficiently and intuitively by a human operator through one teleoperated robotic actuator.

\section{Method}
\subsection{Scenario}
\begin{figure}
    \center\scriptsize
    \input{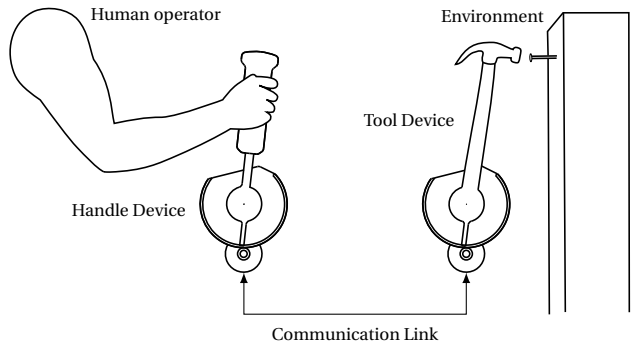}
    \caption{The scenario considered for this study is a human operator interacting with the environment through a bilateral teleoperation system. The teleoperator consists of a handle device and a tool device coupled to each other through a communication link and tuned to mirror velocity and force as accurately as possible.
    The tool device in this experiment was a simulated version of the Dyrac \gls{via} \cite{DyracPaper}.
    }
    \label{fig:variable-impedance-scenario}
\end{figure}
The study is based on a typical bilateral teleoperation setup as illustrated in figure~\ref{fig:variable-impedance-scenario}.
A human operator interacts with the teleoperation system through the handle device, and the teleoperation system interacts with the environment through the tool device.
The teleoperation system mirrors the velocity and force between the handle and the tool device, ideally with full transparency allowing the human operators to interact with the environment as if it was through their own arm.
The tasks to be performed are either a precision task, consisting of moving the tool device from one position to another as fast as possible while stopping precisely at the target position, or a dynamic task, consisting of hitting a target with the tool device with as much impact velocity as possible (cf. section~\ref{sec:variable-impedance-experiment-task}).

In this study, the handle device and tool device are not equal mechanically, but the handle device is rigid whereas the tool device consists of a simulation of the wide impedance range \gls{via} Dyrac \cite{DyracPaper}.
The \gls{via} can be configured with a permanently high impedance setting, permanently low impedance setting, or with a continuously self-adapting impedance (cf. section~\ref{sec:variable-impedance-self-adapting}).
The permanently high impedance setting is equivalent to classical robots and best-suited for precision tasks, when the influence of external forces and end-effector inertia should affect the end-effector position the least possible.
The permanently low impedance setting is equivalent to \glspl{sea} and best-suited for highly dynamic tasks, when impact forces should be reduced and which benefit from energy storage in the actuator's elastic element.
The self-adapting impedance setting uses the proposed ``slow-stiff/fast-soft'' control law to be validated.
It should allow the human operator to perform both types of tasks with comparable performance to the best-suited setting without requiring an explicit mode change.

\subsection{Hypotheses}
The study was based on the following hypotheses:\\
\emph{Hypothesis B1}: A permanently high impedance tool device setting  allows a human operator to perform better in telemanipulated precision tasks than a permanently low impedance setting (according to state of the art robot design).\\[.5em]
\emph{Hypothesis B2}: A permanently low impedance tool device setting allows a human operator to perform better in telemanipulated dynamic tasks than a permanently high impedance setting (according to our previous research \cite{Aiple2017}).\\[.5em]
Hypotheses B1 and B2 are the basic assumptions on which the experiment is designed. Checking them serves as a sanity check for the experiment method and apparatus.\\[.5em]
The hypotheses for validating the new self-adapting impedance variation algorithm were:\\[.5em]
\emph{Hypothesis H1}: A self-adapting control law that increases the tool device impedance at low velocity and decreases its impedance at high velocity allows a human operator to perform telemanipulated precision and dynamic tasks equally well as with the best-suited permanent impedance setting for the respective task while achieving significantly better performance than with the alternative permanent impedance setting for the respective task.\\[.5em]
Hypothesis H1 translates into:\\[.5em]
\hspace*{.05\columnwidth}\begin{minipage}{.93\columnwidth}
    \emph{Hypothesis H1.1}: There is no significant difference in performance achieved in precision tasks with the self-adapting impedance setting or the permanently high impedance setting.\\[.5em]
    \emph{Hypothesis H1.2}: Significantly better performance can be achieved in precision tasks with the self-adapting impedance setting than with the permanently low impedance setting.\\[.5em]
    \emph{Hypothesis H1.3}: Significantly better performance can be achieved in dynamic tasks with the self-adapting impedance setting than with the permanently high impedance setting.\\[.5em]
    \emph{Hypothesis H1.4}: There is no significant difference in performance achieved in dynamic tasks with the self-adapting impedance setting or the permanently low impedance setting.
\end{minipage}

\vspace{0.5em}
\noindent\emph{Hypothesis H2}: The performance achieved with the self-adapting impedance setting does not depend on whether the type of task performed is the same over many repetitions or few repetitions.\\[.5em]
Hypothesis H2 thus suggests that hypotheses H1.1 to H1.4 are also valid if precision and dynamic tasks are performed in quick succession.
Hypotheses H1.1 to H1.4 will be referred to as H2.1 to H2.4 if used for quickly succeeding precision and dynamic tasks.

\subsection{Experiment Apparatus}
\subsubsection{Hardware}
\begin{figure}
    \centering
    \includegraphics[width=7cm]{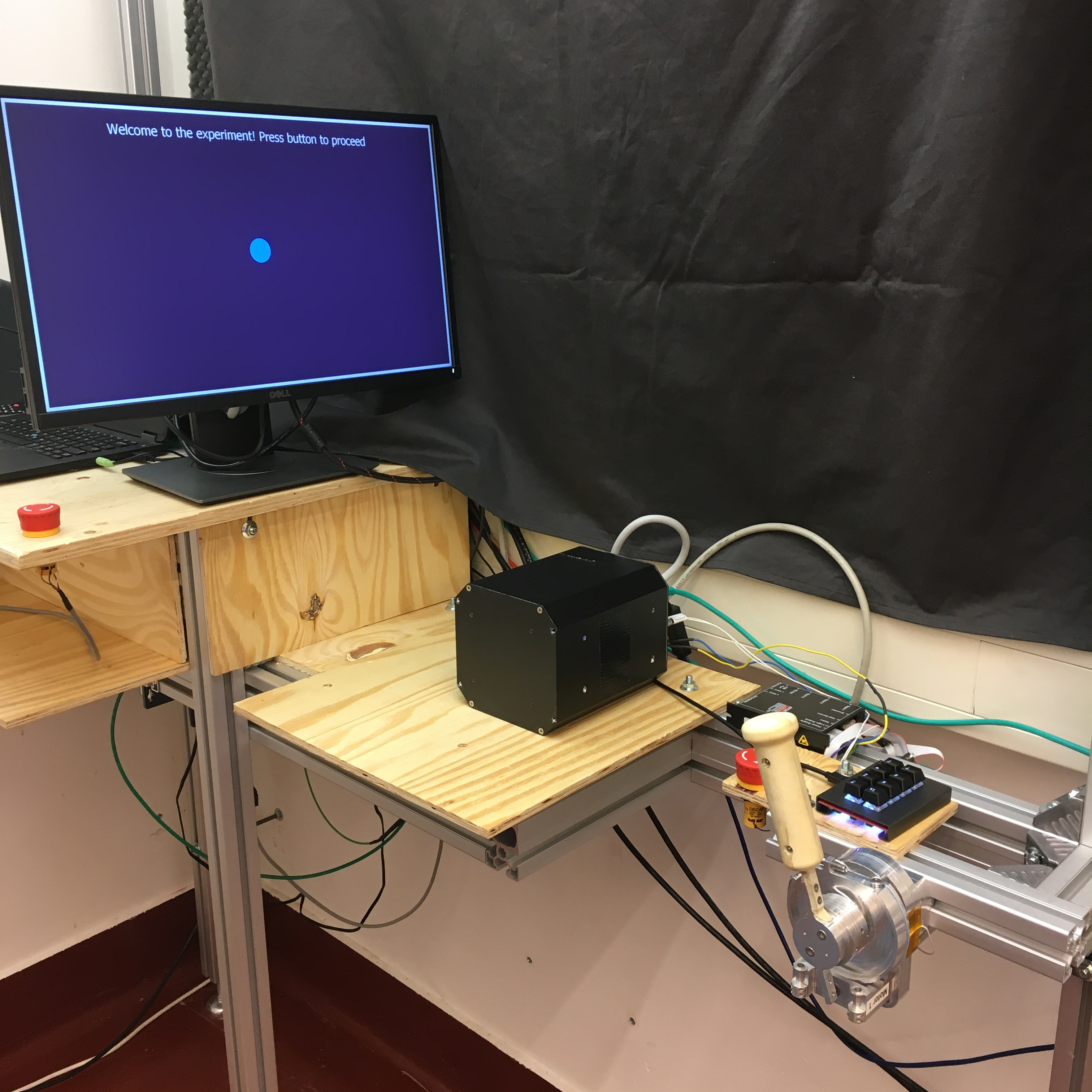}
    \caption{Photo of the setup used for the experiment. The white handle in front is the handle of the 1-DOF haptic joystick, the black box in the center is the control computer running real-time Linux, and the screen visualizes the GUI with the instructions and feedback for the participants (controlled by a second computer not visible on the picture).}
    \label{fig:variable-impedance-photo-setup}
\end{figure}

The experiment apparatus consisted of a haptic bilateral teleoperation setup with one degree of freedom, where the handle device consisted of a stiff actuator device (cf. figure~\ref{fig:variable-impedance-photo-setup}) and the tool device of a simulation model of the Dyrac \gls{via} \cite{DyracPaper}, similar to the architecture proposed by Christiansson \cite{Christiansson2007}, but with a \gls{via} instead of a permanently soft tool device.
A simulation of the actuator was used instead of the physical actuator to make the experiment outcome independent of some flaws that had been detected in the Dyrac prototype \cite{DyracPaper}.
Mainly, considerable hysteresis was measured in the prototype of up to \SI{40}{\percent}.
Further, as the prototype had been built without gears, the position holding accuracy of the joint position motor was deemed insufficient in situations with dynamic external loads, like impacts.

\begin{figure}
    \centering
    \footnotesize
    \input{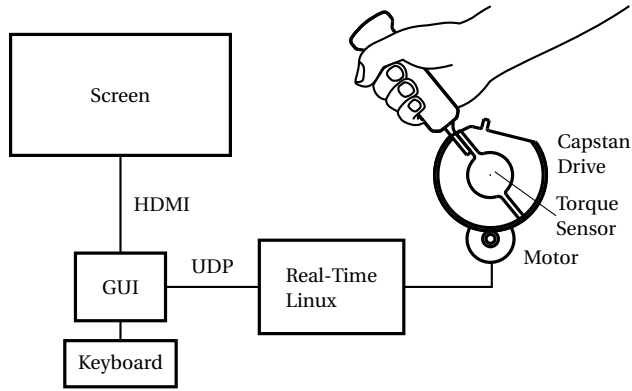}
    \caption{Schematic of the setup used for the experiment}
    \label{fig:variable-impedance-schematic-setup}
\end{figure}

Except for removing hysteresis and position holding inaccuracy under load, the simulation model was built according to the system identification of the Dyrac prototype and included a model of the kinematic of the stiffness changing mechanism (end-effector inertia \SI{0.0125}{\kg\square\m}, joint motor set position to actual position transfer function as second order Butterworth filter with \SI{20}{\Hz} cut-off frequency, stiffness changing motor set position to actual position transfer function as second order Butterworth filter with \SI{10}{\Hz} cut-off frequency).
The simulation model ran at \SI{1}{\kHz} on a real-time Linux computer.
A computer screen connected to a second computer showed the user interface (GUI) to the participants, leading them through the experiment (cf. figure~\ref{fig:variable-impedance-schematic-setup}).
The computer showing the GUI was connected through a UDP connection over Gigabit Ethernet with the real-time Linux computer.

\subsubsection{User Interface}
\label{sec:variable-impedance-user-interface}
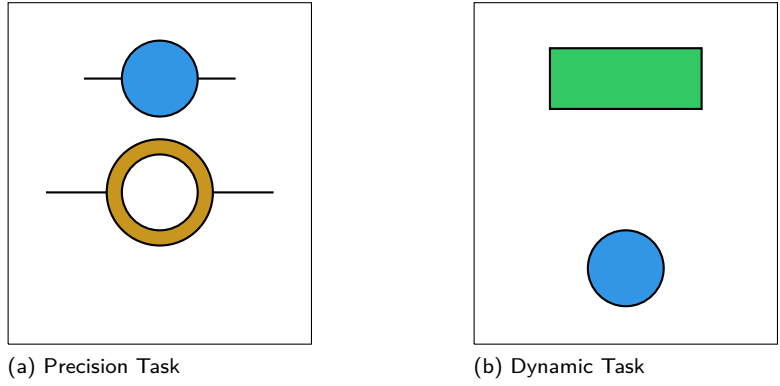
\begin{figure}
    \centering
    \subfloat[Precision Task]{% !TeX TS-program = pdflatex
% !BIB TS-program = biber
% !TeX root = root.tex
% !TeX encoding = UTF-8
% !TeX spellcheck = en_US

\footnotesize
\begin{tikzpicture}
[explanations/.style={font={\ttfamily\scriptsize}},anchor=west,align=left]
	\draw (-2cm,-2.5cm) -- (2cm,-2.5cm) -- (2cm,2cm) -- (-2cm,2cm) -- (-2cm,-2.5cm);
%	\node[font={\bfseries\sffamily\scriptsize},anchor=center] at (0,2cm) {Follow the target!};
	\draw[thick,fill=actualbrush] (0,1cm) circle[radius=0.5cm];
	\draw[thick,fill=targetbrushposition] (0,-0.5cm) circle[radius=0.7cm];
	\draw[thick,fill=white] (0,-0.5cm) circle[radius=0.5cm];
    \draw[thick] (-1.5,-0.5) -- (-0.7,-0.5);
    \draw[thick] (1.5,-0.5) -- (0.7,-0.5);
    \draw[thick] (-1,1) -- (-0.5,1);
    \draw[thick] (1,1) -- (0.5,1);
	
%	\node[explanations] at (-3.8cm,2cm) {Instructions};
%	\node[explanations] at (-3.8cm,1cm) {User cursor};
%	\node[explanations] at (-3.8cm,-0.5cm) {Target};
\end{tikzpicture}}
    \hspace{2cm}
    \subfloat[Dynamic Task]{% !TeX TS-program = pdflatex
% !BIB TS-program = biber
% !TeX root = root.tex
% !TeX encoding = UTF-8
% !TeX spellcheck = en_US

\footnotesize
\begin{tikzpicture}
[explanations/.style={font={\ttfamily\scriptsize}},anchor=west,align=left]
\draw (-2cm,-2.5cm) -- (2cm,-2.5cm) -- (2cm,2cm) -- (-2cm,2cm) -- (-2cm,-2.5cm);
%\node[font={\bfseries\sffamily\scriptsize},anchor=center] at (0,2cm) {Hit the target!};
\draw[thick,fill=actualbrush] (0,-1.5cm) circle[radius=0.5cm];
\draw[thick,fill=boxbrush] (-1,0.6cm) rectangle (1cm,1.4cm);

%\node[explanations] at (-3.8cm,2cm) {Instructions};
%\node[explanations] at (-3.8cm,-1.5cm) {User cursor};
%\node[explanations] at (-3.8cm,0.6cm) {Target};
\end{tikzpicture}}
    \caption{Essential elements of the user interface: the blue circle visualizes the tool device position, the brown circle visualizes the target position for the precision task, and the green box visualizes the impact position for the dynamic task.}
    \label{fig:variable-impedance-ui}
\end{figure}

The user interface was a full screen application (cf. figure~\ref{fig:variable-impedance-photo-setup}) with one line for instructions, a user cursor whose position corresponded to the position of the \gls{via} output position, and a target whose appearance changed depending on the current task (cf. the instruction video of the experiment in the complementary material \cite{Aiple2019b}).
The rotatory position of the handle was mapped linearly to a linear position on the screen.
For the precision task, the target was a moving ring, and the participants were instructed to follow the target as closely as possible, such that the user cursor filled the ring as much as possible at any time (cf. Fig.~\ref{fig:variable-impedance-ui}).
For the dynamic task, the target was a rectangle, and the participants were instructed to hit the target with a speed of the user cursor as high as possible.
After every trial the result achieved in that trial was displayed to the participants in the instruction line.

\subsection{Self-Adapting Teleoperation System Tuning and Identification}
\label{sec:variable-impedance-self-adapting}
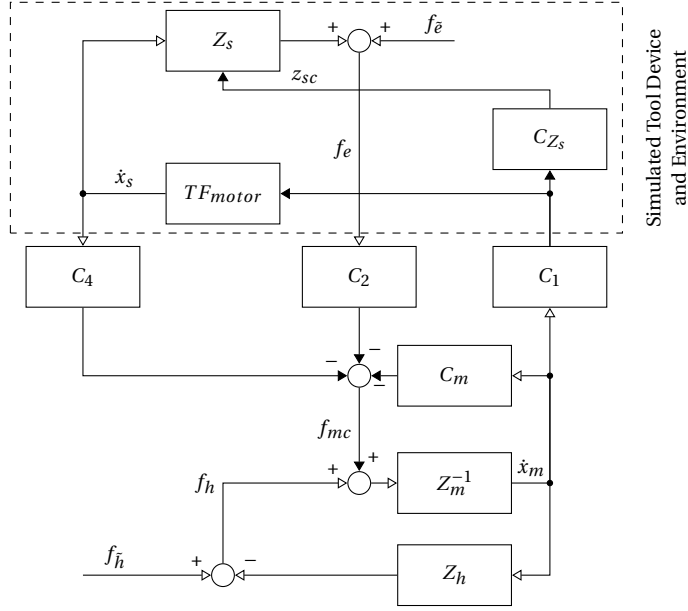
\begin{figure}
    \centering
    \footnotesize
    % !TeX TS-program = lualatex
% !BIB TS-program = bibtex
% !TeX root = controller-architecture-exp3.tex
% !TeX encoding = UTF-8
% !TeX spellcheck = en_US
\begin{tikzpicture}[
sysblock/.style={node distance=1cm,rectangle,draw,minimum width=1.5cm,minimum height=0.8cm},
signal/.style={>=Triangle,->},
measure/.style={>=Triangle[open],->},
sum/.style={draw,circle,minimum size=0.3cm}]
\coordinate (sysref) at (0cm,0cm);

\node[sysblock] (C1) {\Cone};

\node[sysblock] (C2) [left=  of C1]{\Ctwo};

\node (commright) [fit=(C1) (C2) (C1)]{};
\node[sysblock] (Cm) [node distance=0.4cm,below=of commright]{\Cm};

\node[sysblock] (Zminv) [node distance=0.6cm,below=of Cm]{\Zminv};
\node[sysblock] (TFmotor) [node distance=0.4cm,above left=of C2]{\TFmotor};
\draw let \p1 = (C2),\p2 = (TFmotor) in node (C3placeholder) at (\x2,\y1){};
\node[sysblock] (C4) [left=  of C3placeholder]{\Cfour};
\node (comm) [fit=(C1) (C4) (C1)]{};
\draw let \p1 = (Zminv),\p2 = (C2) in node at (\x2,\y1) [sum] (summaster){};
\draw let \p1 = (Cm),\p2 = (C2) in node at (\x2,\y1) [sum] (summastercontrol){};
\node[sysblock] (CZs)  [node distance=1cm,above=of C1] {\CZs};

\node[sysblock] (Zh) [node distance=0.4cm,below=of Zminv] {\Zh};
\node[sysblock] (Zs) [node distance=1.2cm,above=of TFmotor] {\Zs};
\draw let \p1 = (Zh),\p2 = (C3placeholder) in node (sumhext) at (\x2,\y1) [sum] {};
\draw let \p1 = (Zs),\p2 = (C2) in node (sumeext) at (\x2,\y1) [sum] {};

\draw let \p1 = (summaster),\p2 = (C3placeholder) in coordinate (fh) at (\x2,\y1) ;

\draw let \p1 = (Zs),\p2 = (C2) in coordinate (fe) at (\x2,\y1) ;
\draw let \p1 = (Zh),\p2 = (C4) in coordinate (fhext) at (\x2,\y1) ;
\draw let \p1 = (Zs),\p2 = (commright) in coordinate (feext) at (\x2,\y1) ;

\junction{(C4)}{(TFmotor)}
\junction{(C4)}{(TFmotor)}

\junction{(C1)}{(Zminv)}
\junction{(C1)}{(Cm)}
\junction{(C1)}{(TFmotor)}

\draw[signal] (Cm) -- (summastercontrol);% node [midway, below] {\fCm};
\draw[measure] (summaster) -- (Zminv);% node [midway, above] {\fm};
\draw[signal] (C2) -- (summastercontrol);
\draw[signal] (C1) |- (TFmotor);
\draw[signal] (summastercontrol) -- (summaster) node [midway, left] {\fmc};

\draw[measure] (Zminv) -| (C1) node [near start, above] {\vm};% node [near start, below] {\vh};
\draw[measure] (TFmotor) -| (C4) node [near start, above] {\vs};% node [near start, below] {\ve};
\draw[signal] (C4) |- (summastercontrol);
\draw let \p1 = (Zminv),\p2 = (C1) in coordinate (Zminvjunction) at (\x2,\y1);
\draw[measure] (Zminvjunction) |- (Zh);
\draw[measure] (Zminvjunction) |- (Cm);
\draw[measure] (Zh) -- (sumhext);
\draw[measure] (sumhext) |- (summaster) node [midway,left] {\fh};

\draw let \p1 = (TFmotor),\p2 = (C4) in coordinate (TFmotorjunction) at (\x2,\y1);
\draw[measure] (TFmotorjunction) |- (Zs);
\draw[measure] (Zs) -- (sumeext);
\draw[measure] (sumeext) -- (C2) node [midway, left] {\fe};

\draw[measure] (fhext) -- (sumhext) node [near start, above] {\fhext};
\draw[measure] (feext) -- (sumeext) node [near start, above] {\feext};
\draw[signal] (C1) -- (CZs);
\draw let \p1 = (CZs),\p2 = (TFmotor),\p3 = (Zs) in coordinate (Zsset1) at (\x1,{(\y1+\y3)/2});
\draw let \p1 = (CZs),\p2 = (TFmotor),\p3 = (Zs) in coordinate (Zsset2) at (\x2,{(\y1+\y3)/2});
\draw[signal] (CZs) -- (Zsset1) -- (Zsset2)  node [near end,above]{\Zsset} -- (Zs);

\node (plus) [node distance=1pt,above left=of summaster.west]{$+$};
\node (plus) [node distance=1pt,above right=of summaster.north]{$+$};
\node (plus) [node distance=1pt,above left=of summastercontrol.west]{$-$};
\node (plus) [node distance=1pt,above right=of summastercontrol.north]{$-$};

\node (plus) [node distance=1pt,below right=1pt and -2pt of summastercontrol.east]{$-$};

\node (plus) [node distance=1pt,above left=of sumhext.west]{$+$};
\node (plus) [node distance=1pt,above right=of sumhext.east]{$-$};
\node (plus) [node distance=1pt,above left=of sumeext.west]{$+$};
\node (plus) [node distance=1pt,above right=of sumeext.east]{$+$};

\draw[dashed] let \p1=(C4.west),\p2=(C4.north),\p3=(C1.east),\p4=(Zs.north) in ($(\x1,\y2)+(-0.2,0.17)$) rectangle ($(\x3,\y4)+(0.2,0.1)$) node at ($0.5*(\x3,\y2)+0.5*(\x3,\y4)$) [anchor=north,rotate=90,inner sep=0.5cm]{\begin{minipage}{3cm}\centering Simulated Tool Device and Environment\end{minipage}};
\end{tikzpicture}
    \caption{Controller architecture of the teleoperator (Adapted from \cite{Hashtrudi-zaad2001}). The tool device and the environment were simulated. \TFmotor is the transfer function modeling the time behavior of the position controlled joint position motor of the \gls{via}. The tool device impedance \Zs was not constant, but was modified depending on the handle device velocity \vm through the parameter \CZs, as indicated by the signal \Zsset. This implemented the self-adapting impedance variation algorithm. The simulated environment force \feext was zero for the precision task in free air, or rendered a contact at the target position during the dynamic task.
        }
    \label{fig:variable-impedance-controller-architecture}
\end{figure}

Fig.~\ref{fig:variable-impedance-controller-architecture} shows the controller architecture of the teleoperation system.
The control laws \Cm, \Cone, \Ctwo and \Cfour were tuned for optimal transparency according to the rules of Hashtrudi-Zaad and Salcudean~\cite{Hashtrudi-zaad2001}.
The tool device and the environment were simulated.
The tool device was position controlled, so \Cthree from the four-channel architecture was not used.
In the diagram, \TFmotor is the transfer function modeling the time behavior of the position controlled joint position motor of the \gls{via} according to the Dyrac system identification.
The tool device impedance \Zs was not constant, but was modified depending on the handle device velocity \vm through the parameter \CZs, as indicated by the signal \Zsset.
This implemented the self-adapting impedance variation algorithm.
The time behavior of \Zs was also modeled according to the Dyrac system identification.
The simulated environment force \feext was zero for the precision task in free air, or rendered a contact at the target position during the dynamic task ($k_{wall} = \SI{10000}{\N\m\per\radian}$, $b_{wall} = \SI{20}{\N\m\s\per\radian}$).

The control law \CZs could be changed to switch between the low impedance setting (mode L), the high impedance setting (mode H), and the adaptive setting (mode A).
Table~\ref{tab:variable-impedance-controller-gains} shows the controller gains used for the different modes.
The low impedance setting was chosen such that a resonance frequency of \SI{4.5}{\Hz} was achieved ($k = \SI{10}{\N\m\per\radian}$, $b = \SI{0.01}{\N\m\s\per\radian}$).
The high impedance setting corresponded to the highest stable impedance setting of the actuator ($k = \SI{100}{\N\m\per\radian}$, $b = \SI{0.8}{\N\m\s\per\radian}$).
In the adaptive setting, the control law mapped the handle velocity \vm to the tool device impedance.
The mapping law was such that the lowest impedance setting corresponded to the impedance of mode L and the highest impedance setting corresponded to the impedance of mode H.

\begin{table}
    \centering
    \begin{threeparttable}
        \caption{Controller gains of the system shown in~Fig.~\ref{fig:variable-impedance-controller-architecture}}
        \label{tab:variable-impedance-controller-gains}
        \renewcommand*{\arraystretch}{1.6}
        \small
        \begin{tabular}{p{5ex}cccp{6cm}}
            \toprule
            Contr. & \multicolumn{4}{c}{Parameter}\\
            \cmidrule(l){2-5}
            & \Cm = - \Cfour & \Cone  & \Ctwo & \CZs \\
            \midrule
            H &  0.8 + $\frac{10}{\slap}$ & 1 & 1 & $k = \SI{100}{\N\m\per\radian}$, $b = \SI{0.8}{\N\m\s\per\radian}$ \\
            L &  0.8 + $\frac{10}{\slap}$ & 1 & 1 & $k = \SI{10}{\N\m\per\radian}$, $b = \SI{0.01}{\N\m\s\per\radian}$\\
            A &  0.8 + $\frac{10}{\slap}$ & 1 & 1 & $k = f(\vm)$, $b = f(\vm)$\\
            \bottomrule
        \end{tabular}
    \end{threeparttable}
\end{table}

The velocity ranges for the adaptive impedance variation for distinguishing between precision task and dynamic task were based on our previous experiments on teleoperated hammering for the dynamic task \cite{Aiple2018} and were refined in the pilot phase to an upper velocity of a precision task of $\vmmin = \SI{1}{\radian\per\s}$, and a lower velocity of a dynamic task of $\vmmax = \SI{6}{\radian\per\s}$. 
An exponential transition was chosen between these two points in order to map the relatively small velocity range (\SIrange{1}{6}{\radian\per\s}) to the high actuator stiffness range (\SIrange{10}{100}{\N\m\per\radian}).
These constraints resulted in the following equations for the control law:
\begin{dmath}
    \kfunc{\vm} = \min(\kmax, (\kmax - \kmin) \exp(-\avar \frac{(\vm-\vmmin)}{(\vmmax-\vmmin})) + \kmin)
\end{dmath},
\begin{dmath}
    \bfunc{\vm} = \min(2 \, \Dfunc{\vm} \, \sqrt{\kstiffness\mmass}, 0.8)
\end{dmath},
with
\begin{dmath}
    \Dfunc{\vm} = (\Dmax - \Dmin) \exp(-\avar \frac{(\vm-\vmmin)}{(\vmmax-\vmmin)}) + \Dmin
\end{dmath},

and $k_{max} = \SI{100}{\N\m\per\radian}$, $k_{min} = \SI{10}{\N\m\per\radian}$, $D_{max} = \SI{0.7}{}$, $D_{min} = \SI{0.01}{}$, $\avar=\SI{6}{}$,  $\mmass = \SI{0.0125}{\kg\m\squared}$.

\begin{figure}
    \setlength{\figurewidth}{8cm}
    \setlength{\figureheight}{6cm}
    \centering
    \footnotesize
    \input{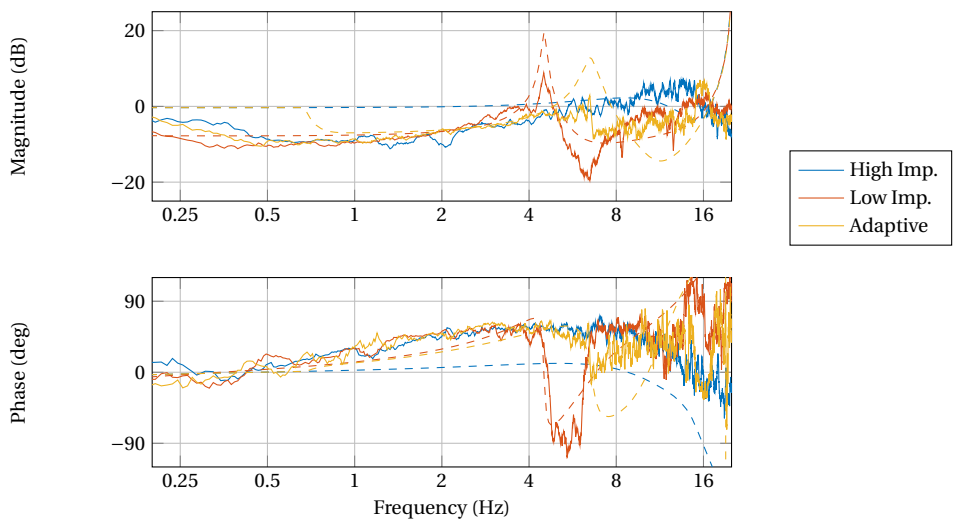}
    \caption{Bode plots of the perceived impedance at the different stiffness settings.
    The dashed lines show the results as expected from the analytical solution of the system behavior.
    The continuous lines show the measured results.
    }
    \label{fig:variable-impedance-ztomodel-bode}
\end{figure}

Figure~\ref{fig:variable-impedance-ztomodel-bode} shows the magnitude over frequency of the resulting transfer function at the different settings as obtained through system identification measurements.
As can be seen, the adaptive setting follows the ``slow-stiff/fast-soft'' principle: the curve corresponding to the adaptive setting is close to that of the high impedance setting at low frequencies and close to that of the low impedance setting at high frequencies.
There is also a narrow region with high gain showing the resonance, although at a higher frequency of approximately \SI{6.5}{\Hz} than the low impedance resonance.

\subsection{Conditions}
Six different conditions were compared in the experiment:
\begin{itemize}
    \item[PR] Precision task reference condition: The participants performed the precision task while the \gls{via} was set to the permanently high impedance setting.
    \item[PC] Precision task counter-reference condition: The participants performed the precision task while the \gls{via} was set to the permanently low impedance setting.
    \item[PA] Precision task adaptive condition: The participants performed the precision task while the \gls{via} was set to the self-adapting impedance setting.
    \item[DR] Dynamic task reference condition: The participants performed the dynamic task while the \gls{via} was set to the permanently low impedance setting.
    \item[DC] Dynamic task counter-reference condition: The participants performed the dynamic task while the \gls{via} was set to the permanently high impedance setting.
    \item[DA] Dynamic task adaptive condition: The participants performed the dynamic task while the \gls{via} was set to the self-adapting impedance setting.
\end{itemize}

Additionally, participants were given the opportunity to familiarize with the device in permanently high impedance (FH condition), low impedance (FL condition), and self-adapting impedance setting (FA condition) before starting the measurements. No specific objective was given to the participants for the familiarization.

\subsection{Participants}
The experiment was performed by 24 participants (16 male, 8 female, age 17 - 64 median 23.5) grouped in four groups of six participants each.
The participants were students and researchers with a higher education background.
None of the participants had prior experience with the experiment. The experiment had been approved by the human research ethics committee of TU Delft and all participants gave written informed consent before participation.

The participants watched a \SI{3}{\minute} instruction video before the start of the experiment explaining the goal and giving instructions \cite{Aiple2019b}.
In previous experiments, the instructions included extensive explanations about resonance of an elastic tool and how to exploit it to achieve higher output velocities \cite{Aiple2018}.
In this experiment, this kind of explanations was left out on purpose.
The motivation was to evaluate whether the participants could reproduce the results of the previous experiments with even less knowledge about the tool device.
The target instructed to the participants was to reach a travel time of less than \SI{300}{\ms} travel time for the precision task and to reach an output velocity of more than \SI{10}{\radian\per\s} for the dynamic task (cf.~section~\ref{sec:variable-impedance-dependent-measures}).
The result per trial was displayed to the participants after each trial (cf.~section~\ref{sec:variable-impedance-user-interface}).

\subsection{Experiment Procedure}
\label{sec:variable-impedance-experiment-procedure}
Fig.~\ref{fig:variable-impedance-experiment-conditions} shows the experiment procedure divided into four phases: I.~familiarization,
II.~training trials,
III.~reference measurement, and 
IV.~ task switch test.
The familiarization phase gave the participants time to familiarize with the experiment device and to the dynamics of the actuator in high impedance (FH), low impedance (FL) and adaptive (FA) configurations.
The training trials and reference measurement were identical except for the target positions (see below), but only the reference measurement was taken into account for the analysis.
Training trials and reference measurement consisted of a series of six blocks with twenty trials per block.
The target positions changed between the trials, such that every second trial was at the same distance of \SI{0.3}{\radian} of the previous trial either forward or backward, whereas the other trials were at varying distance compared to the previous trial.
This was done for three reasons:
\begin{enumerate}
\item the target positions should be spread over the full range of motion in order to compensate for a possible influence of the working point on the outcome (the positions were spread equally from \SIrange{-0.1789}{0.5}{\radian} with \SI{0}{\radian} corresponding to the vertical handle position, positive angles forward; one position at \SI{-0.42}{\radian} was added, which resulted in a pose of the participants with the hand very close to the own body to have a data point with an adverse pose);
\item the next target position should not be easily anticipated in order to avoid a possible influence of repetitive motions on the outcome;
\item the travel distance of some measurements should be equal for the positioning task, in order to make the measurements comparable.
\end{enumerate}
The last two reasons do not apply to the dynamic task, but for simplicity the same target positions were used for both tasks.
However, this means that only ten trials per block were effectively used for analysis of the positioning task (those with a difference of \SI{+-0.3}{\radian} to the previous target position), whereas twenty trials per block were used for analysis of the dynamic task.

The order of positions was the same for all groups and for all trial blocks, but different between phase II and III to prevent prior knowledge of the next target position during phase III.
The order of condition blocks during phases II and III was assigned to the groups according to the Latin square method as shown in Fig.~\ref{fig:variable-impedance-latinsquare}.
Goal of phase IV, the task switch test, was to test how easily the participants could switch from one task to another with the adaptive impedance variation of the tool device.
According to hypothesis H2, we expected phase III and phase IV to have the same outcome.
For phase IV ten blocks of the PA and DA condition (precision or dynamic task with adaptive stiffness) were performed in an order randomized per group, where each block consisted of two trials with the first trial at a varying distance from the zero position and the second trial at \SI{\pm0.3}{\radian} of the previous position.

Before every block, the participants had to bring the handle to the zero position to start.
Before the first trial of every block, the participants had to press a key on a keyboard to indicate that they were ready.
Thus, the participants were able to rest between blocks if they wanted and mentally prepare for the next task.
This was important especially for phase IV, where each block consisted only of two trials, but the type of task could switch from one block to the next.

\begin{figure}
    \centering
    \small
    % !TeX root = root.tex

\footnotesize
\begin{tikzpicture}
[box/.style={draw},
phasebox/.style={anchor=west,minimum height=0.7cm,minimum width=0.7cm,align=left},
legendbox/.style={anchor=west,minimum height=0.7cm,minimum width=0.7cm,align=left},
condbox/.style={box,anchor=west,minimum height=0.7cm,minimum width=0.7cm,align=center}
]
\node[phasebox](familiarizationtitle) at (0,0cm) {\textbf{I. Familiarization:}};
\node[legendbox](legend) at (0,-0.7cm) {Condition: \\ Time:};
\node[condbox](fh) [right=0.5cm of legend]{FH \\ 2 min};
\node[condbox](fl) [right=0cm of fh]{FL \\ 2 min};
\node[condbox](fa) [right=0cm of fl]{FA \\ 2 min};
\node[phasebox](trainingtitle) at (0,-1.5cm) {\textbf{II. Trainig Trials:}};
\node[legendbox](legend) at (0,-2.2cm) {Condition: \\ Trials:};
\node[condbox](t1) [right=0.5cm of legend]{1 \\ 20};
\node[condbox](t2) [right=0cm of t1]{2 \\ 20};
\node[condbox](t3) [right=0cm of t2]{3 \\ 20};
\node[condbox](t4) [right=0cm of t3]{4 \\ 20};
\node[condbox](t5) [right=0cm of t4]{5 \\ 20};
\node[condbox](t6) [right=0cm of t5]{6 \\ 20};
%\draw [decorate,decoration={brace,amplitude=10pt},xshift=0pt,yshift=10pt]
%(pr.north west) -- (dc.north east) node [black,midway,yshift=0.6cm] {Reference};
\node[phasebox](referencetitle) at (0,-3cm){\textbf{III. Reference Measurement:}};
\node[legendbox](legend2) at (0,-3.7cm){Condition: \\ Trials:};
\node[condbox](r1) [right=0.5cm of legend2]{1 \\ 20};
\node[condbox](r2) [right=0cm of r1]{2 \\ 20};
\node[condbox](r3) [right=0cm of r2]{3 \\ 20};
\node[condbox](r4) [right=0cm of r3]{4 \\ 20};
\node[condbox](r5) [right=0cm of r4]{5 \\ 20};
\node[condbox](r6) [right=0cm of r5]{6 \\ 20};
%\draw [decorate,decoration={brace,amplitude=10pt},xshift=0pt,yshift=10pt]
%(pa10.north west) -- (da10.north east) node [black,midway,yshift=0.6cm] {Control Law Test};
\node[phasebox](hditest) at (0,-4.5cm) {\textbf{IV. Task Switch Test:}};
\node[legendbox](legend3) at (0,-5.2cm){Condition: \\ Trials:};
\node[condbox](pa1) [right=0.5cm of legend3]{PA \\ 2};
\node[condbox](da1) [right=0cm of pa1]{DA \\ 2};
\draw [decorate,decoration={brace,amplitude=5pt},xshift=0pt,yshift=-5pt]
(da1.south east) -- (pa1.south west) node [black,midway,yshift=-0.4cm] {10 x in randomized order};
\end{tikzpicture}
    \caption{Order of experiment conditions grouped by objective. The order was shown for participants of group 1. The order of conditions during phase II was altered according to the Latin square method (cf. figure~\ref{fig:variable-impedance-latinsquare}). The order of conditions during phases III and IV was changed for groups 3 and 4 by swapping PA and DA.}
    \label{fig:variable-impedance-experiment-conditions}
\end{figure}

\begin{figure}
    \centering
    \small
    % !TeX TS-program = pdflatex
% !BIB TS-program = biber
% !TeX root = root.tex
% !TeX encoding = UTF-8
% !TeX spellcheck = en_US

% Draw latin square with all combinations of two conditions once
% 312546
% 423651
% 534162
% 645213
% 156324
% 261435
\newcommand{\lsa}{PR}
\newcommand{\lsb}{PC}
\newcommand{\lsc}{PA}
\newcommand{\lsd}{DR}
\newcommand{\lse}{DC}
\newcommand{\lsf}{DA}
\begin{tikzpicture}
[box/.style={draw},
condbox/.style={box,anchor=west,minimum height=0.7cm,minimum width=0.7cm,align=center,node distance=0,inner sep=0,outer sep=0},
ca/.style={condbox,fill=printer1},
cb/.style={condbox,fill=printer2},
cc/.style={condbox,fill=printer3},
cd/.style={condbox,fill=printer4},
ce/.style={condbox,fill=printer5},
cf/.style={condbox,fill=printer6}]

\node[condbox](c) at (0,0){};
\node[condbox](t1) [right=of c]{1};
\node[condbox](t2) [right=of t1]{2};
\node[condbox](t3) [right=of t2]{3};
\node[condbox](t4) [right=of t3]{4};
\node[condbox](t5) [right=of t4]{5};
\node[condbox](t6) [right=of t5]{6};

\node(lt) [above right=0.1cm and 1.75cm of c,anchor=south,node distance=0,inner sep=0,outer sep=0]{\textbf{Trial Block}};
\node(lg) [rotate=90,below left=1.75cm and 0.1cm of c,anchor=south,node distance=0,inner sep=0,outer sep=0]{\textbf{Group}};

\node[condbox](ga) [below=of c]{A};
\node[condbox](gb) [below=of ga]{B};
\node[condbox](gc) [below=of gb]{C};
\node[condbox](gd) [below=of gc]{D};
\node[condbox](ge) [below=of gd]{E};
\node[condbox](gf) [below=of ge]{F};

\node[cc](a1) [right=of ga]{\lsc};
\node[cd](b1) [right=of gb]{\lsd};
\node[ce](c1) [right=of gc]{\lse};
\node[cf](d1) [right=of gd]{\lsf};
\node[ca](e1) [right=of ge]{\lsa};
\node[cb](f1) [right=of gf]{\lsb};

\node[ca](a2) [right=of a1]{\lsa};
\node[cb](b2) [right=of b1]{\lsb};
\node[cc](c2) [right=of c1]{\lsc};
\node[cd](d2) [right=of d1]{\lsd};
\node[ce](e2) [right=of e1]{\lse};
\node[cf](f2) [right=of f1]{\lsf};

\node[cb](a3) [right=of a2]{\lsb};
\node[cc](b3) [right=of b2]{\lsc};
\node[cd](c3) [right=of c2]{\lsd};
\node[ce](d3) [right=of d2]{\lse};
\node[cf](e3) [right=of e2]{\lsf};
\node[ca](f3) [right=of f2]{\lsa};

\node[ce](a4) [right=of a3]{\lse};
\node[cf](b4) [right=of b3]{\lsf};
\node[ca](c4) [right=of c3]{\lsa};
\node[cb](d4) [right=of d3]{\lsb};
\node[cc](e4) [right=of e3]{\lsc};
\node[cd](f4) [right=of f3]{\lsd};

\node[cd](a5) [right=of a4]{\lsd};
\node[ce](b5) [right=of b4]{\lse};
\node[cf](c5) [right=of c4]{\lsf};
\node[ca](d5) [right=of d4]{\lsa};
\node[cb](e5) [right=of e4]{\lsb};
\node[cc](f5) [right=of f4]{\lsc};

\node[cf](a6) [right=of a5]{\lsf};
\node[ca](b6) [right=of b5]{\lsa};
\node[cb](c6) [right=of c5]{\lsb};
\node[cc](d6) [right=of d5]{\lsc};
\node[cd](e6) [right=of e5]{\lsd};
\node[ce](f6) [right=of f5]{\lse};
\end{tikzpicture}
    \caption{Latin square showing the order of conditions in the trial blocks (1-6) per participant group (A-F): PA = position--adaptive, PR = position--reference (high impedance), PC = position--counter-reference (low impedance), DA = dynamic--adaptive, DR = dynamic--reference (low impedance), DC = dynamic--counter-reference (high impedance).}
    \label{fig:variable-impedance-latinsquare}
\end{figure}

\subsection{Experiment Task}
\label{sec:variable-impedance-experiment-task}
In both tasks, the instructions for a new block of trials were shown to the participants until they indicated that they were ready by pressing a key on the keyboard.
For the precision task, the target cursor then appeared every \SI{2}{\s} at a new position and the participants had to follow the cursor to its new position.
For the dynamic task, the participants moved to a new starting position before every trial, located close below the new target position, while the target box was still invisible to the participants.
Once the participants had reached the starting position, the target appeared and the participants could perform one hammering strike.
The starting position was on purpose close to the target position, to insinuate the participants to perform the hammering motion as a backward-forward strike instead of trying to hit the target in a straight forward motion.
A hammering trial was over when the target had been reached by the tool device.
This was indicated to the participants by changing the color of the target from green to orange.
In both tasks, data were collected during the trial and the result was shown on the screen to the participants immediately after each trial.

\subsection{Dependent Measures}
\label{sec:variable-impedance-dependent-measures}
All relevant data were captured at \SI{1}{\kHz} directly from the sensors integrated in the experiment apparatus and logged for later analysis.
Task performance was calculated on-the-fly in the control system and communicated via UDP to the user interface for display.
Table~\ref{tab:variable-impedance-dependent-measures} lists the dependent measures used in the analysis of the experiment outcome.
Travel time and integral of time-weighted absolute error (ITAE) were chosen as metrics for the precision task, similar to the settling time and ITAE known from control theory as indicators for the step-response positioning performance of a controller through a plant.
For this, the travel time was defined as settling time minus dead time, calculated from the moment that the participants start moving to the moment that the tool position settled at the target position.
This method of calculation was chosen instead of the classical settling time in order to compensate for effects of different reaction times of the participants.
The threshold for moving was defined as \SI{0.03}{\radian} from the handle position when the target position jumped to its new value.
Settling was defined as maintaining an absolute error from the target position of less than \SI{0.03}{\radian} for a duration of \SI{0.5}{\s}.
The travel time was calculated during each trial and displayed to the participants at the end of each trial.

The ITAE is an indicator of the controller performance favoring fast initial response even at the cost of slightly overshooting.
It was added as pilot participants often could not reliably judge whether they had already reached the target position corridor, resulting in very long calculated travel times even for very small residual errors just above the settling threshold.
The ITAE metric corresponds more to the intuition of an operator where a small remaining error or small oscillations around the target position are considered as less critical than a slow initial response.
It was calculated in the post-processing analysis as 
\begin{dmath}
    ITAE = \int_0^T t \left|x-x_0\right| dt
\end{dmath}
with $t$ the time variable, $T$ the travel time, $x$ the tool position, and $x_0$ the target position \cite{Seborg2016}.

For the dynamic task, the maximum output velocity was calculated and displayed to the participants as the peak forward velocity of the tool device before impact.
Furthermore, the velocity gain was calculated in the post-processing analysis as the ratio of maximum output velocity over maximum handle velocity (as done in \cite{Aiple2018}).
The maximum output velocity measures how effective the strike was, as a higher velocity means more energy on impact, whereas the gain measures how efficient the strike was, as a higher gain means that more energy from the backwards motion could be recovered into forward motion through the elastic actuator.

\begin{table}
    \caption{Dependent Measures of the experiment for the precision task and the dynamic task}
    \label{tab:variable-impedance-dependent-measures}
    \centering
    \small
    \begin{threeparttable}
        \begin{tabular}{cccc}
            \toprule
            Precision Task & & Dynamic Task &\\
            \midrule
            Travel time\tnote{$\bigstar$} & \SI{}{\s} & Maximum output velocity\tnote{$\bigstar$} & \SI{}{\radian\per\s} \\
            ITAE\tnote{$\blacklozenge$}        & \SI{}{\radian\s\squared} & Gain                    &      ---                \\
            \bottomrule
        \end{tabular}
        \begin{tablenotes}
            \item [$\bigstar$] Result shown to participants immediately after every trial
            \item [$\blacklozenge$] Integral of time-weighted absolute error
        \end{tablenotes}
    \end{threeparttable}
\end{table}

\subsection{Statistical Evaluation}
As explained in section~\ref{sec:variable-impedance-experiment-procedure}, 10 values were obtained per trial block per participant for the precision task and 20 values were obtained per trial block per participant for the dynamic task.
Wilcoxon's signed-rank test \cite{Wilcoxon1945} was used for testing the experiment hypotheses, as it allows pairwise testing of related samples without assuming normal distribution of the data.
Assuming that samples are related compensates for per-participant effects on the results.
And the data could not be assumed to be normally distributed as it was always more difficult to achieve ``good'' performance (low travel time, low ITAE, high output velocity or high gain) than ``bad'' performance.

The comparison of the results of low impedance setting and high impedance setting served to verify the basic assumptions B1 and B2.
Further, the following pairs of settings were compaired: adaptive impedance -- high impedance, and adaptive impedance -- low impedance.
The test was applied to the results of the precision task and the dynamic task, corresponding to the four hypotheses H1.1-H1.4.

H1.1 and H1.4 (no significant difference in performance between adaptive setting and reference setting results) are supported if the null-hypothesis of Wilcoxon's signed-rank test could not be rejected at the \SI{5}{\percent}-level.
H1.2 and H1.3 (significantly better performance of adaptive setting results than counter-reference setting results) are supported if the null-hypothesis of Wilcoxon's signed-rank test could be rejected at a level better than \SI{5}{\percent}.
The same tests were performed for the results of the task switch test phase to verify hypothesis 2 stating that hypotheses H1.1 to H1.4 (then referred to as H2.1 to H2.4) also hold for quick succession of precision and dynamic tasks.
These results are labeled with ``Adaptive*'' or ``A*'' for short in the result tables and figures.

The statistical tests were applied once on the per-participant medians per condition and once on per-participant very good trials per condition.
The per-participant \percentiles{10} per condition were considered as measure for very good trials for the precision task (lower is better for travel time and ITAE measures).
The per-participant \percentiles{90} per condition were considered as measure for very good trials for the dynamic task (higher is better for maximum output velocity and gain measures).
With 10 trials, the \nth{10} or \percentile{90} corresponds to the middle between the best and the second best trials.
With 20 trials, the \nth{10} or \percentile{90} corresponds to the middle between the second best and the third trial.
Using the \nth{10} or \percentiles{90} instead of the minima or maxima adds some basic robustness to outliers to the evaluation.
Comparing also the very good trials of the participants in addition to the median of the trials gives an impression of how much effort was required from the participants to obtain a certain result with a specific impedance setting (cf. the discussion section~\ref{sec:variable-impedance-discussion}).

\section{Results}
The experiment measurement data are available in the supplementary material \cite{Aiple2020b}.

\subsection{Precision task}

\begin{figure*}
    \setlength{\figurewidth}{5cm}
    \setlength{\figureheight}{4cm}
    \footnotesize
    \centering
    \subfloat[High Impedance]{\input{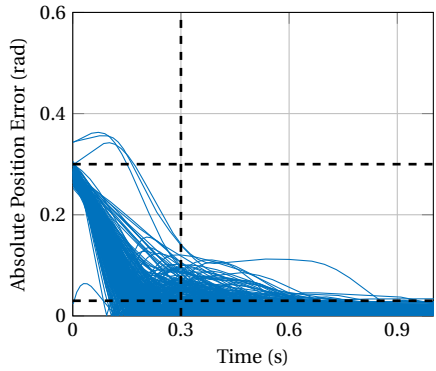}}
    \hspace{1cm}
    \subfloat[Low Impedance]{\input{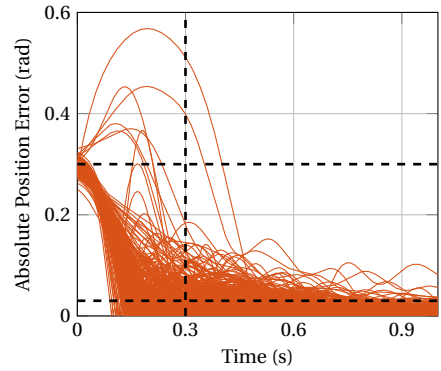}}
    \\[1cm]
    \subfloat[Adaptive]{\input{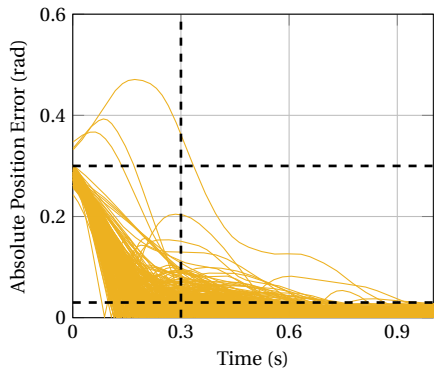}}
    \hspace{1cm}
    \subfloat[Adaptive*]{\input{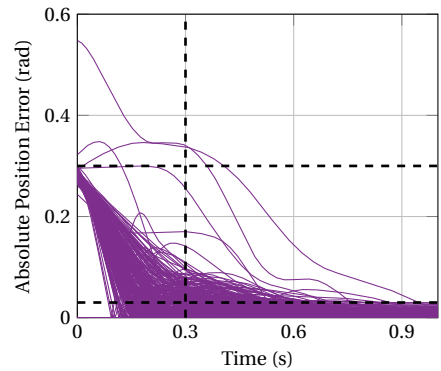}}    
    \caption{Precision task position error plot of the reference measurements of all participants for trials in high, low and adaptive settings, and of the task switch test for trials in the adaptive* setting. The dashed lines show the experiment criteria: the initial absolute error was \SI{0.3}{\radian}, and the goal communicated to the participants was to achieve a travel time of less than \SI{0.3}{\s}. The criterion for the end of the travel time was to keep the absolute error under \SI{0.03}{\radian} for at least \SI{0.5}{\s}.}
    \label{fig:variable-impedance-positionplots}
\end{figure*}

\begin{figure}\RawFloats
\captionsetup[table]{position=top}
    \setlength{\figurewidth}{8cm}
    \setlength{\figureheight}{5cm}
    \footnotesize
    \centering
\colorlet{mycolor1}{matlab1}%
\colorlet{mycolor2}{matlab2}%
\colorlet{mycolor3}{matlab3}%
\colorlet{mycolor4}{matlab4}%
\begin{tikzpicture}

\begin{axis}[%
width=0.951\figurewidth,
height=\figureheight,
at={(0\figurewidth,0\figureheight)},
scale only axis,
xmin=0,
xmax=25,
xtick={ 1,  7, 13, 19, 24},
xlabel={Participant (-)},
ymin=0,
ymax=1.5,
ymajorgrids,
ylabel={Travel Time (\si{\s})},
legend style={legend cell align=left, align=left, draw=black}
]
\draw[dashed, line width=0.1pt] (axis cs:0,0.3) -- (axis cs:25,0.3);
\addplot [color=mycolor1, draw=none, mark=*, mark options={solid, fill=mycolor1, mycolor1}]
 plot [error bars/.cd, y dir = both, y explicit]
 table[row sep=crcr, y error plus index=2, y error minus index=3]{%
1	0.3975	0.2675	0.1425\\
2	0.3755	0.2805	0.2165\\
3	0.1795	0.1175	0.0225\\
4	0.2305	0.2285	0.1165\\
5	0.2355	0.2635	0.0625\\
6	0.4085	0.1675	0.2265\\
7	0.294	0.166	0.125\\
8	0.438	0.242	0.269\\
9	0.345	0.202	0.132\\
10	0.3055	0.4885	0.1075\\
11	0.339	0.172	0.179\\
12	0.2385	0.2775	0.0445\\
13	0.242	0.285	0.058\\
14	0.4445	0.1795	0.2425\\
15	0.3285	0.3655	0.0995\\
16	0.4765	0.1775	0.2615\\
17	0.278	0.243	0.047\\
18	0.369	0.257	0.129\\
19	0.375	0.063	0.215\\
20	0.4245	0.2325	0.2415\\
21	0.299	0.235	0.111\\
22	0.33	0.305	0.135\\
23	0.4405	0.0955	0.0615\\
24	0.383	0.077	0.094\\
};
\addlegendentry{High Imp. Median}

\addplot [color=mycolor2, draw=none, mark=*, mark options={solid, fill=mycolor2, mycolor2}]
 plot [error bars/.cd, y dir = both, y explicit]
 table[row sep=crcr, y error plus index=2, y error minus index=3]{%
1	0.659	0.434	0.198\\
2	0.5745	0.2565	0.2275\\
3	0.342	0.176	0.164\\
4	0.584	0.635	0.312\\
5	0.486	0.348	0.256\\
6	0.623	0.313	0.421\\
7	0.4165	0.1265	0.1095\\
8	0.3825	0.1915	0.2155\\
9	0.507	0.331	0.143\\
10	0.361	0.186	0.083\\
11	0.445	0.17	0.319\\
12	0.3155	0.3475	0.1465\\
13	0.498	0.189	0.216\\
14	0.6385	0.1255	0.5005\\
15	0.533	0.174	0.288\\
16	0.2855	0.3075	0.1155\\
17	0.589	0.27	0.233\\
18	0.5135	0.1175	0.3595\\
19	0.64	0.102	0.138\\
20	0.4155	0.2145	0.2295\\
21	0.504	0.115	0.15\\
22	0.456	0.227	0.281\\
23	0.524	0.171	0.197\\
24	0.332	0.179	0.145\\
};
\addlegendentry{Low Imp. Median}

\addplot [color=mycolor3, draw=none, mark=*, mark options={solid, fill=mycolor3, mycolor3}]
 plot [error bars/.cd, y dir = both, y explicit]
 table[row sep=crcr, y error plus index=2, y error minus index=3]{%
1	0.4705	0.2215	0.1845\\
2	0.213	0.292	0.056\\
3	0.324	0.175	0.117\\
4	0.459	0.054	0.274\\
5	0.3475	0.1235	0.1805\\
6	0.445	0.163	0.199\\
7	0.3355	0.2075	0.1575\\
8	0.281	0.237	0.091\\
9	0.392	0.239	0.195\\
10	0.4545	0.0895	0.2475\\
11	0.341	0.231	0.17\\
12	0.2205	0.1135	0.0305\\
13	0.417	0.174	0.227\\
14	0.521	0.191	0.341\\
15	0.4525	0.2345	0.2255\\
16	0.276	0.302	0.069\\
17	0.3385	0.2185	0.1285\\
18	0.474	0.168	0.197\\
19	0.2595	0.1835	0.1345\\
20	0.2205	0.3045	0.0355\\
21	0.306	0.101	0.111\\
22	0.263	0.21	0.102\\
23	0.494	0.143	0.168\\
24	0.447	0.148	0.146\\
};
\addlegendentry{Adaptive Median}

\addplot [color=mycolor4, draw=none, mark=*, mark options={solid, fill=mycolor4, mycolor4}]
 plot [error bars/.cd, y dir = both, y explicit]
 table[row sep=crcr, y error plus index=2, y error minus index=3]{%
1	0.4635	0.3335	0.3505\\
2	0.285	0.3	0.146\\
3	0.1995	0.2925	0.0265\\
4	0.256	0.213	0.1\\
5	0.325	0.121	0.135\\
6	0.4695	0.1515	0.3095\\
7	0.1995	0.2955	0.0635\\
8	0.2825	0.3045	0.0985\\
9	0.4225	0.1535	0.1235\\
10	0.3085	0.2365	0.1355\\
11	0.2645	0.2775	0.1295\\
12	0.353	0.166	0.156\\
13	0.436	0.078	0.287\\
14	0.271	0.215	0.124\\
15	0.23	0.622	0.00900000000000001\\
16	0.3055	0.3075	0.1165\\
17	0.326	0.169	0.162\\
18	0.455	0.076	0.271\\
19	0.4075	0.0855	0.2765\\
20	0.302	0.314	0.137\\
21	0.23	0.197	0.038\\
22	0.371	0.211	0.195\\
23	0.3755	0.0855	0.0915\\
24	0.3795	0.2295	0.1215\\
};
\addlegendentry{Adaptive* Median}
\draw[->,>=latex] (0.05\figurewidth,0.9\figureheight) -- (0.05\figurewidth,0.8\figureheight) node [midway,anchor=west] {Lower is better};
\end{axis}
\end{tikzpicture}%
    \caption{Per-participant median results of travel time measured during the precision task with \SI{95}{\percent} confidence intervals (lower is better).
    It was communicated to the participants that they should try to achieve a travel time of less than \SI{0.3}{\s}.
    This is indicated by the dashed line.
    The expected outcome according to the experiment hypotheses was that the shortest travel times would be achieved with the high impedance setting (blue), the longest with the low impedance (red), and that the results with the adaptive (yellow) and adaptive* (purple) settings would not be significantly different from the results with the high impedance settings.
    }
    \label{fig:variable-impedance-participants-traveltime-median}
\vspace{2cm}
    \captionof{table}{Grouped results over all participants of pairwise comparison of per-participant median travel times measured during the precision task, given as effect size of median difference and \SI{95}{\percent} confidence interval (negative difference indicates better performance of first setting compared to second, $n=24$, $p$-values refer to Wilcoxon's signed-rank test results).
    The last column indicates which experiment hypothesis was checked by the statistical test and whether the outcome supports (green) or contradicts (red) the hypothesis (significance level $\alpha = \SI{5}{\percent}$).
    }
    \label{tab:variable-impedance-results-traveltime-median}
    \small
    % !TeX TS-program = lualatex
% !BIB TS-program = bibtex
% !TeX root = tab-results-traveltime-median.tex
% !TeX encoding = UTF-8
% !TeX spellcheck = en_US
% Changelog:

\begin{tabular}{cL{0.11\columnwidth}cC{0.15\columnwidth}C{0.22\columnwidth}C{0.1\columnwidth}C{0.06\columnwidth}}
        \toprule
        \multicolumn{3}{c}{Impedance Setting Pair} & Median Diff. & \CI{} & $p$ & Hyp.\\
        \multicolumn{3}{c}{} & (\si{\ms}) & (\si{\ms}) & & \\
        \midrule
        \legendRectangle{LowImp} & Low -- High & \legendRectangle{HighImp} & 162 & [106, 205] & < 0.001 & \colorbox{colorHypConf}{B1} \\
        \midrule
        \legendRectangle{Adaptive} & Adaptive -- High & \legendRectangle{HighImp} & 50 & [2, 77] & 0.29 & \colorbox{colorHypConf}{H1.1} \\
        \legendRectangle{Adaptive} & Adaptive -- Low & \legendRectangle{LowImp} & -110 & [-178, -81] & < 0.001 & \colorbox{colorHypConf}{H1.2} \\
        \midrule
        \legendRectangle{Taskswitch} & Adaptive* -- High & \legendRectangle{HighImp} & 12 & [-75, 48] & 0.56 & \colorbox{colorHypConf}{H2.1} \\
        \legendRectangle{Taskswitch} & Adaptive* -- Low & \legendRectangle{LowImp} & -151 & [-217, -85] & < 0.001 & \colorbox{colorHypConf}{H2.2} \\
        \bottomrule
    \end{tabular}
\end{figure}

\begin{figure}\RawFloats
\captionsetup[table]{position=top}
    \setlength{\figurewidth}{8cm}
    \setlength{\figureheight}{5cm}
    \footnotesize
    \centering
\colorlet{mycolor1}{matlab1}%
\colorlet{mycolor2}{matlab2}%
\colorlet{mycolor3}{matlab3}%
\colorlet{mycolor4}{matlab4}%
\begin{tikzpicture}

\begin{axis}[%
width=0.951\figurewidth,
height=\figureheight,
at={(0\figurewidth,0\figureheight)},
scale only axis,
xmin=0,
xmax=25,
xtick={ 1, 7, 13, 19, 24},
xlabel={Participant (-)},
ymin=0,
ymax=0.038,
ymajorgrids,
ylabel={ITAE (\si{\radian\square\s})},
legend style={legend cell align=left, align=left}
]
\addplot [color=mycolor1, draw=none, mark=*, mark options={solid, fill=mycolor1, mycolor1}]
 plot [error bars/.cd, y dir = both, y explicit]
 table[row sep=crcr, y error plus index=2, y error minus index=3]{%
1	0.00543398210433886	0.00229171062929234	0.00345363785321528\\
2	0.00324601498148039	0.00772958290843298	0.0019203898200417\\
3	0.00169259032252684	0.00074940959651764	0.000350015442973068\\
4	0.0019997222530765	0.00336383756126303	0.00123835180074875\\
5	0.00277800519776651	0.00290578811710766	0.00102387164696428\\
6	0.00446967385701152	0.00414613394437994	0.00259136328889035\\
7	0.0028789120903916	0.003099598253836	0.00124705752251999\\
8	0.00593732197055835	0.00448130514325423	0.0043296763966256\\
9	0.00459784470101877	0.00645680914517134	0.00252971856298824\\
10	0.00444529703751056	0.00672986147292491	0.00228375747255653\\
11	0.00307749627364551	0.00440005573488743	0.00179909813982083\\
12	0.00310169922604023	0.00606137442063952	0.000911675233237015\\
13	0.00278031518757595	0.00523479825022147	0.000852081493015303\\
14	0.00504539853059497	0.00277710549496015	0.00289287288968102\\
15	0.00406780674608728	0.0100577016533633	0.00127040596193492\\
16	0.00705499534908583	0.00658452271448405	0.00441840102753826\\
17	0.00369142327633022	0.00358516339232855	0.000640233069368995\\
18	0.00417433939220031	0.00537301357868512	0.00135054495415249\\
19	0.0029276019321928	0.00252357006383163	0.00146498833165026\\
20	0.00420729045361212	0.00644155230752681	0.00204402721580608\\
21	0.00271355694496994	0.00730404657740207	0.000950896202226547\\
22	0.00397569199367136	0.00950464593187868	0.00188681270389225\\
23	0.00973692256277303	0.0056167644350947	0.00178010667224164\\
24	0.0052768637653997	0.00103762197579835	0.00159840118245339\\
};
\addlegendentry{High Imp. Median}

\addplot [color=mycolor2, draw=none, mark=*, mark options={solid, fill=mycolor2, mycolor2}]
 plot [error bars/.cd, y dir = both, y explicit]
 table[row sep=crcr, y error plus index=2, y error minus index=3]{%
1	0.0107206603037366	0.0157592921549922	0.00510120009144509\\
2	0.00649236317968346	0.00976942903186647	0.00434588511694797\\
3	0.00302442261538726	0.00713585733465774	0.00133734483031646\\
4	0.010082638039507	0.00993833010621331	0.00804890658102007\\
5	0.00496067554574336	0.00664410069190921	0.00297226688041681\\
6	0.00946805944603491	0.00893482560218741	0.00731161934281252\\
7	0.00345471294640376	0.00664076630975803	0.00121437313997367\\
8	0.00402000651277656	0.00187911328378136	0.0022736885548946\\
9	0.00585847542228754	0.00808477118138049	0.00229929594260178\\
10	0.0038154713539438	0.00693861843313247	0.00112792429551708\\
11	0.00422113332769495	0.00445722162001198	0.00332634286109053\\
12	0.00323034912514946	0.00717668399304887	0.00158997762727987\\
13	0.00612822059021849	0.00555215491344426	0.00432306833291278\\
14	0.00884997789796552	0.0065539224745729	0.00774468659994232\\
15	0.00841066269126364	0.00586843075852492	0.00538816308812005\\
16	0.00327230776537318	0.00695719422272783	0.00165111261478814\\
17	0.00842683706812276	0.00840871509710841	0.00494830374370735\\
18	0.00499862996391808	0.00788162320719801	0.00356516142554377\\
19	0.00979588981787387	0.00622242985372235	0.00383555710923932\\
20	0.00416304536174756	0.00265552703905546	0.00238502351295014\\
21	0.00414287263247224	0.00611754123781202	0.00107270137636614\\
22	0.0046563633301677	0.00491525091772672	0.00350487448082831\\
23	0.0138245324145557	0.0050384620794955	0.00781364258935562\\
24	0.00384649196962371	0.00387099101294837	0.00231496170880519\\
};
\addlegendentry{Low Imp. Median}

\addplot [color=mycolor3, draw=none, mark=*, mark options={solid, fill=mycolor3, mycolor3}]
 plot [error bars/.cd, y dir = both, y explicit]
 table[row sep=crcr, y error plus index=2, y error minus index=3]{%
1	0.00757606153074901	0.0124174897263837	0.00470226611809404\\
2	0.0022744410413607	0.0067208153300679	0.000753293980233488\\
3	0.00293542364591951	0.00268940738519462	0.00084967169861107\\
4	0.00497315056751849	0.00197407009697433	0.00359489081740056\\
5	0.00298011040177312	0.00190824660637654	0.0013986896568613\\
6	0.00488791355482212	0.00531859935552421	0.00203737106296866\\
7	0.00330643017401032	0.00375032267758197	0.0012728260682578\\
8	0.00270618555468282	0.00390314943315195	0.000703896346917949\\
9	0.00496402910469924	0.00806296824732041	0.00352390185761431\\
10	0.00583857246786168	0.00500429471186979	0.00330838071262757\\
11	0.00324979057148186	0.00373582491148234	0.00131788338784387\\
12	0.00271951048486762	0.000810405838579391	0.000686633809870961\\
13	0.00545419184024696	0.00354595841453806	0.00318725527531097\\
14	0.00555791530562817	0.00636126251033944	0.00374254990793418\\
15	0.00539947105540861	0.00605949502787188	0.00253362844596416\\
16	0.00432444175115476	0.00466702342976415	0.00177604955051145\\
17	0.0031765967772478	0.00322030535655004	0.000556676022309385\\
18	0.00552790547593646	0.0105097701721156	0.00222952278729841\\
19	0.00231882227329245	0.00255941866402991	0.0013907963489045\\
20	0.00257312887798346	0.00514887744515378	0.000526671251884258\\
21	0.00305824616622534	0.00165483644153759	0.000967109547333958\\
22	0.00274481428367945	0.00261660354017187	0.0012585183346525\\
23	0.0102882699412922	0.00441786368266791	0.00380468650472768\\
24	0.00629622520872343	0.00303422529016521	0.00221799789339186\\
};
\addlegendentry{Adaptive Median}

\addplot [color=mycolor4, draw=none, mark=*, mark options={solid, fill=mycolor4, mycolor4}]
 plot [error bars/.cd, y dir = both, y explicit]
 table[row sep=crcr, y error plus index=2, y error minus index=3]{%
1	0.00573819556298833	0.0094080804824003	0.00499897684754892\\
2	0.00372528772767639	0.00411492764890844	0.00264687647521806\\
3	0.00217497338781354	0.00304042318787297	0.000454413877841955\\
4	0.00194619998049859	0.00363866039220319	0.000654162112619093\\
5	0.00394649544782388	0.0036529767650717	0.0019967894592987\\
6	0.00813959870593855	0.00498536615775617	0.00670573342688159\\
7	0.00222946336980584	0.00534101292264801	0.00120185329692129\\
8	0.00302192872408419	0.00960046253826057	0.0011053735463061\\
9	0.00767932680135895	0.00296347931154426	0.00346035068747981\\
10	0.00417796592016123	0.00327025816792352	0.00247124733017431\\
11	0.00204595711797796	0.00505171473418082	0.000979589755757647\\
12	0.00364592517190061	0.00487447266500572	0.00127310097512798\\
13	0.00414813045880219	0.00314285747056949	0.00284413981815829\\
14	0.00217110582354265	0.00574559914294689	0.000912673554437431\\
15	0.00314133985374153	0.0136290666006593	0.000331221664736335\\
16	0.00476004169141338	0.00669301875533067	0.00259009347894866\\
17	0.00299519282322171	0.00422054057378655	0.00150954583365429\\
18	0.00603872047637658	0.00119213446858107	0.00395516014755375\\
19	0.00398876456637859	0.00311451265711218	0.00302748178274864\\
20	0.00337729712664907	0.00745063143806258	0.00185788768365164\\
21	0.00291825820458105	0.00297437364873466	0.000863330710513204\\
22	0.00378442934699542	0.00444731391546034	0.00221807966541025\\
23	0.00732789129423118	0.00440341469297807	0.00260727768856759\\
24	0.00709226226744531	0.00281909309117957	0.00348558435350118\\
};
\addlegendentry{Adaptive* Median}
\draw[->,>=latex] (0.05\figurewidth,0.9\figureheight) -- (0.05\figurewidth,0.8\figureheight) node [midway,anchor=west] {Lower is better};
\end{axis}
\end{tikzpicture}%
    \caption{Per-participant median results of ITAE measured during the precision task with \SI{95}{\percent} confidence intervals (lower is better).
    The expected outcome according to the experiment hypotheses was that the lowest ITAE would be achieved with the high impedance setting (blue), the highest with the low impedance (red), and that the results with the adaptive (yellow) and adaptive* (purple) settings would not be significantly different from the results with the high impedance settings.
    }
    \label{fig:variable-impedance-participants-itae-median}
\vspace{2cm}
    \captionof{table}{Grouped results over all participants of pairwise comparison of per-participant median ITAEs measured during the precision task, given as effect size of median difference and \SI{95}{\percent} confidence interval (negative difference indicates better performance of first setting compared to second, $n=24$, $p$-values refer to Wilcoxon's signed-rank test results).
    The last column indicates which experiment hypothesis was checked by the statistical test and whether the outcome supports (green) or contradicts (red) the hypothesis (significance level $\alpha = \SI{5}{\percent}$).
    }
    \label{tab:variable-impedance-results-itae-median}
    \small
    % !TeX TS-program = lualatex
% !BIB TS-program = bibtex
% !TeX root = tab-results-itae-median.tex
% !TeX encoding = UTF-8
% !TeX spellcheck = en_US
% Changelog:

\begin{tabular}{cL{0.11\columnwidth}cC{0.15\columnwidth}C{0.22\columnwidth}C{0.1\columnwidth}C{0.06\columnwidth}}
        \toprule
        \multicolumn{3}{c}{Impedance Setting Pair} & Median Diff. & \CI{} & $p$ & Hyp.\\
        \multicolumn{3}{c}{} & (\si{\milli\radian\square\s}) & (\si{\milli\radian\square\s}) & & \\
        \midrule
        \legendRectangle{LowImp} & Low -- High & \legendRectangle{HighImp} & 1.4 & [0.68, 3.8] & 0.0016 & \colorbox{colorHypConf}{B1} \\
        \midrule
        \legendRectangle{Adaptive} & Adaptive -- High & \legendRectangle{HighImp} & 0.39 & [-0.38, 1.0] & 0.34 & \colorbox{colorHypConf}{H1.1} \\
        \legendRectangle{Adaptive} & Adaptive -- Low & \legendRectangle{LowImp} & -1.5 & [-3.1, -0.67] & 0.0011 & \colorbox{colorHypConf}{H1.2} \\
        \midrule
        \legendRectangle{Taskswitch} & Adaptive* -- High & \legendRectangle{HighImp} & 0.076 & [-0.69, 0.54] & 0.88 & \colorbox{colorHypConf}{H2.1} \\
        \legendRectangle{Taskswitch} & Adaptive* -- Low & \legendRectangle{LowImp} & -1.2 & [-2.8, -0.85] & 0.0043 & \colorbox{colorHypConf}{H2.2} \\
        \bottomrule
    \end{tabular}
\end{figure}

\begin{figure}\RawFloats
\captionsetup[table]{position=top}
    \setlength{\figurewidth}{8cm}
    \setlength{\figureheight}{5cm}
    \footnotesize
    \centering
\colorlet{mycolor1}{matlab1}%
\colorlet{mycolor2}{matlab2}%
\colorlet{mycolor3}{matlab3}%
\colorlet{mycolor4}{matlab4}%
\begin{tikzpicture}

\begin{axis}[%
width=0.951\figurewidth,
height=\figureheight,
at={(0\figurewidth,0\figureheight)},
scale only axis,
xmin=0,
xmax=25,
xtick={ 1,  7, 13, 19, 24},
xlabel={Participant (-)},
ymin=0,
ymax=0.8,
ymajorgrids,
ylabel={Travel Time (\si{\s})},
legend style={legend cell align=left, align=left}
]
\draw[dashed, line width=0.1pt] (axis cs:0,0.3) -- (axis cs:25,0.3);
\addplot [color=mycolor1, draw=none, mark=*, mark options={solid, fill=mycolor1, mycolor1}]
 plot [error bars/.cd, y dir = both, y explicit]
 table[row sep=crcr, y error plus index=2, y error minus index=3]{%
1	0.2485	-0.00650000000000001	0.00650000000000001\\
2	0.1495	-0.00950000000000001	0.00950000000000001\\
3	0.1535	-0.0035	0.0035\\
4	0.114	0	0\\
5	0.166	-0.00700000000000001	0.00700000000000001\\
6	0.1695	-0.0125	0.0125\\
7	0.1585	-0.0105	0.0105\\
8	0.16	-0.00900000000000001	0.00900000000000001\\
9	0.208	-0.005	0.005\\
10	0.1915	-0.00650000000000001	0.00650000000000001\\
11	0.15	-0.01	0.01\\
12	0.192	-0.002	0.002\\
13	0.179	-0.00499999999999998	0.00499999999999998\\
14	0.1995	-0.0025	0.0025\\
15	0.2105	-0.0185	0.0185\\
16	0.2095	-0.0055	0.0055\\
17	0.188	-0.043	0.043\\
18	0.222	-0.018	0.018\\
19	0.148	-0.012	0.012\\
20	0.1725	-0.0105	0.0105\\
21	0.1695	-0.0185	0.0185\\
22	0.1935	-0.0015	0.0015\\
23	0.365	-0.014	0.014\\
24	0.252	-0.037	0.037\\
};
\addlegendentry{High Imp. P10}

\addplot [color=mycolor2, draw=none, mark=*, mark options={solid, fill=mycolor2, mycolor2}]
 plot [error bars/.cd, y dir = both, y explicit]
 table[row sep=crcr, y error plus index=2, y error minus index=3]{%
1	0.4225	-0.0385	0.0385\\
2	0.3015	-0.0455	0.0455\\
3	0.172	-0.00599999999999998	0.00599999999999998\\
4	0.269	-0.003	0.003\\
5	0.206	-0.024	0.024\\
6	0.1565	-0.0455	0.0455\\
7	0.253	-0.054	0.054\\
8	0.1625	-0.0045	0.0045\\
9	0.3135	-0.0505	0.0505\\
10	0.239	-0.039	0.039\\
11	0.1215	-0.00449999999999999	0.00449999999999999\\
12	0.16	-0.00900000000000001	0.00900000000000001\\
13	0.2795	-0.0025	0.0025\\
14	0.135	-0.003	0.003\\
15	0.2205	-0.0245	0.0245\\
16	0.1615	-0.00850000000000001	0.00850000000000001\\
17	0.345	-0.011	0.011\\
18	0.1535	-0.0005	0.0005\\
19	0.471	-0.031	0.031\\
20	0.171	-0.015	0.015\\
21	0.274	-0.08	0.08\\
22	0.1545	-0.0205	0.0205\\
23	0.3215	-0.0055	0.0055\\
24	0.1855	-0.0015	0.0015\\
};
\addlegendentry{Low Imp. P10}

\addplot [color=mycolor3, draw=none, mark=*, mark options={solid, fill=mycolor3, mycolor3}]
 plot [error bars/.cd, y dir = both, y explicit]
 table[row sep=crcr, y error plus index=2, y error minus index=3]{%
1	0.209	-0.077	0.077\\
2	0.1485	-0.00850000000000001	0.00850000000000001\\
3	0.1995	-0.00750000000000001	0.00750000000000001\\
4	0.1725	-0.0125	0.0125\\
5	0.1585	-0.00850000000000001	0.00850000000000001\\
6	0.2225	-0.0235	0.0235\\
7	0.1725	-0.00549999999999998	0.00549999999999998\\
8	0.18	-0.00999999999999998	0.00999999999999998\\
9	0.18	-0.017	0.017\\
10	0.2035	-0.0035	0.0035\\
11	0.1505	-0.0205	0.0205\\
12	0.1835	-0.00650000000000001	0.00650000000000001\\
13	0.18	-0.00999999999999998	0.00999999999999998\\
14	0.171	-0.00899999999999998	0.00899999999999998\\
15	0.2185	-0.00850000000000001	0.00850000000000001\\
16	0.207	0	0\\
17	0.2065	-0.0035	0.0035\\
18	0.225	-0.052	0.052\\
19	0.124	-0.001	0.001\\
20	0.1805	-0.0045	0.0045\\
21	0.18	-0.015	0.015\\
22	0.1475	-0.0135	0.0135\\
23	0.325	-0.001	0.001\\
24	0.2835	-0.0175	0.0175\\
};
\addlegendentry{Adaptive P10}

\addplot [color=mycolor4, draw=none, mark=*, mark options={solid, fill=mycolor4, mycolor4}]
 plot [error bars/.cd, y dir = both, y explicit]
 table[row sep=crcr, y error plus index=2, y error minus index=3]{%
1	0.0565	-0.0565	0.0565\\
2	0.0695	-0.0695	0.0695\\
3	0.0865	-0.0865	0.0865\\
4	0.078	-0.078	0.078\\
5	0.095	-0.095	0.095\\
6	0.08	-0.08	0.08\\
7	0.068	-0.068	0.068\\
8	0.092	-0.092	0.092\\
9	0.1495	-0.1495	0.1495\\
10	0.0865	-0.0865	0.0865\\
11	0.0675	-0.0675	0.0675\\
12	0.0985	-0.0985	0.0985\\
13	0.0745	-0.0745	0.0745\\
14	0.0735	-0.0735	0.0735\\
15	0.1105	-0.1105	0.1105\\
16	0.0945	-0.0945	0.0945\\
17	0.082	-0.082	0.082\\
18	0.092	-0.092	0.092\\
19	0.0655	-0.0655	0.0655\\
20	0.0825	-0.0825	0.0825\\
21	0.096	-0.096	0.096\\
22	0.088	-0.088	0.088\\
23	0.142	-0.142	0.142\\
24	0.129	-0.129	0.129\\
};
\addlegendentry{Adaptive* P10}
\draw[->,>=latex] (0.05\figurewidth,0.9\figureheight) -- (0.05\figurewidth,0.8\figureheight) node [midway,anchor=west] {Lower is better};
\end{axis}
\end{tikzpicture}%
    \caption{Per-participant \percentile{10} results of travel time measured during the precision task with \SI{95}{\percent} confidence intervals (lower is better).
    It was communicated to the participants that they should try to achieve a travel time of less than \SI{0.3}{\s}.
    This is indicated by the dashed line.
    The expected outcome according to the experiment hypotheses was that the shortest travel times would be achieved with the high impedance setting (blue), the longest with the low impedance (red), and that the results with the adaptive (yellow) and adaptive* (purple) settings would not be significantly different from the results with the high impedance settings.
    }
    \label{fig:variable-impedance-participants-traveltime-best}
\vspace{2cm}
    \captionof{table}{Grouped results over all participants of pairwise comparison of per-participant \percentile{10} travel times measured during the precision task, given as effect size of median difference and \SI{95}{\percent} confidence interval (negative difference indicates better performance of first setting compared to second, $n=24$, $p$-values refer to Wilcoxon's signed-rank test results).
    The last column indicates which experiment hypothesis was checked by the statistical test and whether the outcome supports (green) or contradicts (red) the hypothesis (significance level $\alpha = \SI{5}{\percent}$).
    }
    \label{tab:variable-impedance-results-traveltime-best}
    \small
    % !TeX TS-program = lualatex
% !BIB TS-program = bibtex
% !TeX root = tab-results-traveltime-best.tex
% !TeX encoding = UTF-8
% !TeX spellcheck = en_US
% Changelog:

\begin{tabular}{cL{0.11\columnwidth}cC{0.15\columnwidth}C{0.2\columnwidth}C{0.1\columnwidth}C{0.06\columnwidth}}
        \toprule
        \multicolumn{3}{c}{Impedance Setting Pair} & Median Diff. & \CI{} & $p$ & Hyp.\\
        \multicolumn{3}{c}{} & (\si{\ms}) & (\si{\ms}) & & \\
        \midrule
        \legendRectangle{LowImp} & Low -- High & \legendRectangle{HighImp} & 14 & [-29, 101] & 0.089 & \colorbox{colorHypRej}{B1} \\
        \midrule
        \legendRectangle{Adaptive} & Adaptive -- High & \legendRectangle{HighImp} & 2 & [-8, 12] & 0.58 & \colorbox{colorHypConf}{H1.1} \\
        \legendRectangle{Adaptive} & Adaptive -- Low & \legendRectangle{LowImp} & -5 & [-94, 24] & 0.14 & \colorbox{colorHypRej}{H1.2} \\
        \midrule
        \legendRectangle{Taskswitch} & Adaptive* -- High & \legendRectangle{HighImp} & -92 & [-106, -83] & < 0.001 & \colorbox{colorHypRej}{H2.1} \\
        \legendRectangle{Taskswitch} & Adaptive* -- Low & \legendRectangle{LowImp} & -111 & [-180, -71] & < 0.001 & \colorbox{colorHypConf}{H2.2} \\
        \bottomrule
    \end{tabular}
\end{figure}

\begin{figure}\RawFloats
\captionsetup[table]{position=top}
    \setlength{\figurewidth}{8cm}
    \setlength{\figureheight}{5cm}
    \footnotesize
    \centering
\colorlet{mycolor1}{matlab1}%
\colorlet{mycolor2}{matlab2}%
\colorlet{mycolor3}{matlab3}%
\colorlet{mycolor4}{matlab4}%
\begin{tikzpicture}

\begin{axis}[%
width=0.951\figurewidth,
height=\figureheight,
at={(0\figurewidth,0\figureheight)},
scale only axis,
xmin=0,
xmax=25,
xtick={ 1,  7, 13, 19, 24},
xlabel={Participant (-)},
ymin=0,
ymax=0.013,
ymajorgrids,
ylabel={ITAE (\si{\radian\square\s})},
legend style={legend cell align=left, align=left}
]
\addplot [color=mycolor1, draw=none, mark=*, mark options={solid, fill=mycolor1, mycolor1}]
 plot [error bars/.cd, y dir = both, y explicit]
 table[row sep=crcr, y error plus index=2, y error minus index=3]{%
1	0.00188650629473748	-9.38379563860954e-05	9.38379563860954e-05\\
2	0.00121887893029941	-0.000106746231139279	0.000106746231139279\\
3	0.00129154989826979	-5.1024981283979e-05	5.1024981283979e-05\\
4	0.000747413318482777	-1.39571338449816e-05	1.39571338449816e-05\\
5	0.00157566102629249	-0.000178472524509743	0.000178472524509743\\
6	0.00169170805722587	-0.0001866025108953	0.0001866025108953\\
7	0.00148349006244866	-0.000148364505422953	0.000148364505422953\\
8	0.0014712480368983	-0.00013639753703445	0.00013639753703445\\
9	0.00173619074349842	-0.000331935394532111	0.000331935394532111\\
10	0.00210884320189674	-5.26963630572935e-05	5.26963630572935e-05\\
11	0.00116652392258285	-0.000111874211241818	0.000111874211241818\\
12	0.00210694127117291	-8.30827216303054e-05	8.30827216303054e-05\\
13	0.00176118266463475	-0.00016705102992589	0.00016705102992589\\
14	0.00210207858998435	-5.0447050929602e-05	5.0447050929602e-05\\
15	0.00250600860433266	-0.000291392179819695	0.000291392179819695\\
16	0.00261846681681209	-1.81275047354767e-05	1.81275047354767e-05\\
17	0.00210448148304789	-0.000946708723913332	0.000946708723913332\\
18	0.0025657205517712	-0.000258073886276614	0.000258073886276614\\
19	0.00125964717390829	-0.000202966426634242	0.000202966426634242\\
20	0.00187955858268301	-0.000283704655123028	0.000283704655123028\\
21	0.00153988441656141	-0.000222776326181988	0.000222776326181988\\
22	0.00200292648451363	-8.59528052654785e-05	8.59528052654785e-05\\
23	0.00746079617275781	-0.00049601971777358	0.00049601971777358\\
24	0.00326318052658868	-0.000415282056357631	0.000415282056357631\\
};
\addlegendentry{High Imp. P10}

\addplot [color=mycolor2, draw=none, mark=*, mark options={solid, fill=mycolor2, mycolor2}]
 plot [error bars/.cd, y dir = both, y explicit]
 table[row sep=crcr, y error plus index=2, y error minus index=3]{%
1	0.00519437960987492	-0.000425080602416623	0.000425080602416623\\
2	0.00210339701377258	-4.3081048962914e-05	4.3081048962914e-05\\
3	0.00168496656315183	-2.11122191896059e-06	2.11122191896059e-06\\
4	0.00173559828966683	-0.000298133168820059	0.000298133168820059\\
5	0.0018646732492706	-0.000123735416055952	0.000123735416055952\\
6	0.00145883614723335	-0.000697603955989041	0.000697603955989041\\
7	0.00177297815590774	-0.000467361650522351	0.000467361650522351\\
8	0.00165300372908914	-9.33142287928175e-05	9.33142287928175e-05\\
9	0.00349460614117548	-6.45733385102776e-05	6.45733385102776e-05\\
10	0.00225033114778778	-0.000437215910638932	0.000437215910638932\\
11	0.00088041046489561	-1.43800017088123e-05	1.43800017088123e-05\\
12	0.0015271175808255	-0.000113253917044091	0.000113253917044091\\
13	0.00176077071946961	-4.43815378361044e-05	4.43815378361044e-05\\
14	0.00110383294327723	-1.45835474596933e-06	1.45835474596933e-06\\
15	0.00236806647020635	-0.000654433132937237	0.000654433132937237\\
16	0.00145391811926127	-0.000167277031323764	0.000167277031323764\\
17	0.0031345592400668	-0.000343974084348605	0.000343974084348605\\
18	0.00143083886642403	-2.62967195027524e-06	2.62967195027524e-06\\
19	0.00476020486790639	-0.00120012784072816	0.00120012784072816\\
20	0.00161118266813188	-0.000166839180665547	0.000166839180665547\\
21	0.00243104636224243	-0.000639124893863668	0.000639124893863668\\
22	0.00107727950541344	-7.42093439259506e-05	7.42093439259506e-05\\
23	0.00553929996305468	-0.000471589862145369	0.000471589862145369\\
24	0.00153035631508731	-1.17394573121549e-06	1.17394573121549e-06\\
};
\addlegendentry{Low Imp. P10}

\addplot [color=mycolor3, draw=none, mark=*, mark options={solid, fill=mycolor3, mycolor3}]
 plot [error bars/.cd, y dir = both, y explicit]
 table[row sep=crcr, y error plus index=2, y error minus index=3]{%
1	0.00196913381009322	-0.000904661602561753	0.000904661602561753\\
2	0.00130611274550772	-0.000215034315619493	0.000215034315619493\\
3	0.00207557237357748	-1.01795737309602e-05	1.01795737309602e-05\\
4	0.00123379659407762	-0.000144463156040319	0.000144463156040319\\
5	0.00145534281576011	-0.00012607792915171	0.00012607792915171\\
6	0.0025622625949044	-0.000288279896949059	0.000288279896949059\\
7	0.00176586777262768	-0.000267736333124841	0.000267736333124841\\
8	0.00185684048476383	-0.000145448723001037	0.000145448723001037\\
9	0.00127833833432325	-0.000161788912761675	0.000161788912761675\\
10	0.00248401068142142	-4.61810738126888e-05	4.61810738126888e-05\\
11	0.00146278096640332	-0.000469126217234677	0.000469126217234677\\
12	0.00190520567404019	-0.000127671000956462	0.000127671000956462\\
13	0.00199242985991805	-0.000274506705017944	0.000274506705017944\\
14	0.00162267269066903	-0.00019269270702496	0.00019269270702496\\
15	0.00259984237912568	-0.000266000230318776	0.000266000230318776\\
16	0.00250432857113352	-4.40636295097807e-05	4.40636295097807e-05\\
17	0.00250482340386079	-0.000115097351077634	0.000115097351077634\\
18	0.00255701159145144	-0.000741371097186603	0.000741371097186603\\
19	0.000902907202100145	-2.51187222877978e-05	2.51187222877978e-05\\
20	0.00190512057438284	-0.00014133705171636	0.00014133705171636\\
21	0.00184920605319178	-0.000241930565699608	0.000241930565699608\\
22	0.00125196367505376	-0.000234332273973194	0.000234332273973194\\
23	0.00606480219850496	-0.000418781238059569	0.000418781238059569\\
24	0.0039471170196328	-0.000131110295698766	0.000131110295698766\\
};
\addlegendentry{Adaptive P10}

\addplot [color=mycolor4, draw=none, mark=*, mark options={solid, fill=mycolor4, mycolor4}]
 plot [error bars/.cd, y dir = both, y explicit]
 table[row sep=crcr, y error plus index=2, y error minus index=3]{%
1	0.000369609357719704	-0.000369609357719704	0.000369609357719704\\
2	0.000539205626229164	-0.000539205626229164	0.000539205626229164\\
3	0.000860279754985794	-0.000860279754985794	0.000860279754985794\\
4	0.00064601893393975	-0.00064601893393975	0.00064601893393975\\
5	0.000974852994262591	-0.000974852994262591	0.000974852994262591\\
6	0.000716932639528479	-0.000716932639528479	0.000716932639528479\\
7	0.000513805036442274	-0.000513805036442274	0.000513805036442274\\
8	0.000958277588889044	-0.000958277588889044	0.000958277588889044\\
9	0.00210948805693957	-0.00210948805693957	0.00210948805693957\\
10	0.000853359294993464	-0.000853359294993464	0.000853359294993464\\
11	0.000533183681110155	-0.000533183681110155	0.000533183681110155\\
12	0.00118641209838631	-0.00118641209838631	0.00118641209838631\\
13	0.000651995320321946	-0.000651995320321946	0.000651995320321946\\
14	0.000629216134552609	-0.000629216134552609	0.000629216134552609\\
15	0.0014050590945026	-0.0014050590945026	0.0014050590945026\\
16	0.00108497410623236	-0.00108497410623236	0.00108497410623236\\
17	0.000742823494783711	-0.000742823494783711	0.000742823494783711\\
18	0.00104178016441141	-0.00104178016441141	0.00104178016441141\\
19	0.000480641391814975	-0.000480641391814975	0.000480641391814975\\
20	0.000759704721498717	-0.000759704721498717	0.000759704721498717\\
21	0.00102746374703392	-0.00102746374703392	0.00102746374703392\\
22	0.000783174840792586	-0.000783174840792586	0.000783174840792586\\
23	0.00236030680283179	-0.00236030680283179	0.00236030680283179\\
24	0.00180333895697207	-0.00180333895697207	0.00180333895697207\\
};
\addlegendentry{Adaptive* P10}
\draw[->,>=latex] (0.05\figurewidth,0.9\figureheight) -- (0.05\figurewidth,0.8\figureheight) node [midway,anchor=west] {Lower is better};
\end{axis}
\end{tikzpicture}%
    \caption{Per-participant \percentile{10} results of ITAE measured during the precision task with \SI{95}{\percent} confidence intervals (lower is better).
    The expected outcome according to the experiment hypotheses was that the shortest travel times would be achieved with the high impedance setting (blue), the longest with the low impedance (red), and that the results with the adaptive (yellow) and adaptive* (purple) settings would not be significantly different from the results with the high impedance settings.
    }
    \label{fig:variable-impedance-participants-itae-best}
\vspace{2cm}
    \captionof{table}{Grouped results of pairwise comparison of per-participant \percentile{10} ITAEs measured during the precision task, given as effect size of median difference and \SI{95}{\percent} confidence interval (negative difference indicates better performance of first setting compared to second, $n=24$, $p$-values refer to Wilcoxon's signed-rank test results).
    The last column indicates which experiment hypothesis was checked by the statistical test and whether the outcome supports (green) or contradicts (red) the hypothesis (significance level $\alpha = \SI{5}{\percent}$).
    }
    \label{tab:variable-impedance-results-itae-best}
    \small
    % !TeX TS-program = lualatex
% !BIB TS-program = bibtex
% !TeX root = tab-results-itae-best.tex
% !TeX encoding = UTF-8
% !TeX spellcheck = en_US
% Changelog:

\begin{tabular}{cL{0.11\columnwidth}cC{0.15\columnwidth}C{0.22\columnwidth}C{0.1\columnwidth}C{0.06\columnwidth}}
        \toprule
        \multicolumn{3}{c}{Impedance Setting Pair} & Median Diff. & \CI{} & $p$ & Hyp.\\
        \multicolumn{3}{c}{} & (\si{\milli\radian\square\s}) & (\si{\milli\radian\square\s}) & & \\
        \midrule
        \legendRectangle{LowImp} & Low -- High & \legendRectangle{HighImp} & 0.071 & [-0.29, 0.39] & 0.81 & \colorbox{colorHypRej}{B1} \\
        \midrule
        \legendRectangle{Adaptive} & Adaptive -- High & \legendRectangle{HighImp} & 0.091 & [-0.11, 0.31] & 0.33 & \colorbox{colorHypConf}{H1.1} \\
        \legendRectangle{Adaptive} & Adaptive -- Low & \legendRectangle{LowImp} & 0.23 & [-0.41, 0.39] & 0.73 & \colorbox{colorHypRej}{H1.2} \\
        \midrule
        \legendRectangle{Taskswitch} & Adaptive* -- High & \legendRectangle{HighImp} & -1.0 & [-1.3, -0.68] & < 0.001 & \colorbox{colorHypRej}{H2.1} \\
        \legendRectangle{Taskswitch} & Adaptive* -- Low & \legendRectangle{LowImp} & -0.93 & [-1.4, -0.70] & < 0.001 & \colorbox{colorHypConf}{H2.2} \\
        \bottomrule
    \end{tabular}
\end{figure}

Fig.~\ref{fig:variable-impedance-positionplots} shows the plots of the absolute position error during the reference measurements and the task switch test of all participants.
It gives a qualitative comparison between the three settings of the reference measurement plus the task switch test (Adaptive*).
As can be seen in the plots for the low impedance setting, in many cases there are more oscillations and it takes more time to settle below the threshold of \SI{0.03}{\radian} compared to the high and adaptive impedance settings.

Fig.~\ref{fig:variable-impedance-participants-traveltime-median} to \ref{fig:variable-impedance-participants-itae-best} show the per-participant median and \percentile{10} results of the measured travel time and ITAE with \SI{95}{\percent} confidence intervals (lower is better).
The expected outcome according to the experiment hypotheses was that the lowest value of travel time and ITAE would be achieved with the high impedance setting (blue), the highest with the low impedance (red), and that the results with the adaptive (yellow) and adaptive* (purple) settings would not be significantly different from the results with the high impedance settings.

The grouped statistical comparison over all participants of travel time medians (cf. table \ref{tab:variable-impedance-results-traveltime-median}) showed a significant difference at the \siglevel{0.1} between low and high impedance settings (supporting hypothesis B1).
It showed no significant difference at the \siglevel{20} between adaptive and high impedance settings, and adaptive* and high impedance settings, but a significant difference at the \siglevel{0.1} between adaptive and low impedance settings and adaptive* and low impedance settings (supporting hypotheses H1.1, H1.2, H2.1 and H2.2).

The grouped statistical comparison over all participants of ITAE medians (cf. table \ref{tab:variable-impedance-results-itae-median}) showed a significant difference at the \siglevel{0.5} between low and high impedance settings (supporting hypothesis B1).
It showed no significant difference at the \siglevel{20} between adaptive and high impedance settings, and adaptive* and high impedance settings, but a significant difference at the \siglevel{0.5} between adaptive and low impedance settings and adaptive* and low impedance settings (supporting hypotheses H1.1, H1.2, H2.1 and H2.2).

The grouped statistical comparison over all participants of travel time \percentiles{10} (cf. table \ref{tab:variable-impedance-results-traveltime-best}) showed no significant difference at the \siglevel{5} between low and high impedance setting (contradicting hypothesis B1).
It showed no significant difference at the \siglevel{20} between adaptive and high impedance settings, and at the \siglevel{10} between adaptive and low impedance settings, but a significant difference at the \SI{0.1}{\percent}-level between the adaptive* and high impedance settings and adaptive* and low impedance settings (supporting hypotheses H1.2 and H2.2, but contradicting hypotheses H1.1 and H2.1).

The grouped statistical comparison over all participants of ITAE \percentiles{10} (cf. table \ref{tab:variable-impedance-results-itae-best}) showed no significant difference at the \siglevel{20} between low and high impedance settings (contradicting hypothesis B1).
It showed no significant difference at the \siglevel{20} between adaptive and high impedance settings, and adaptive and low impedance settings (supporting hypothesis H1.1, but contradicting hypothesis H1.2), but a significant difference at the \SI{0.1}{\percent}-level between adaptive* and high impedance settings and adaptive* and low impedance settings (supporting hypotheses H1.1 and H2.2, but contradicting hypotheses H1.2 and H2.1).

\subsection{Dynamic task}

\begin{figure*}
    \setlength{\figurewidth}{5cm}
    \setlength{\figureheight}{4cm}
    \footnotesize
    \centering
    \subfloat[High Impedance]{\input{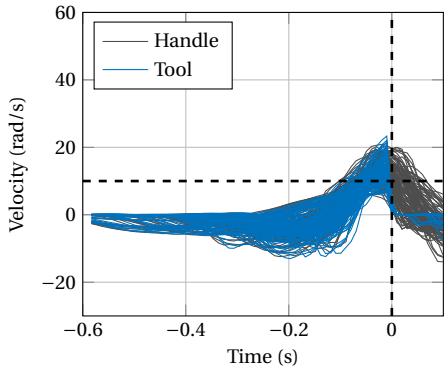}}
    \hspace{1cm}
    \subfloat[Low Impedance]{\input{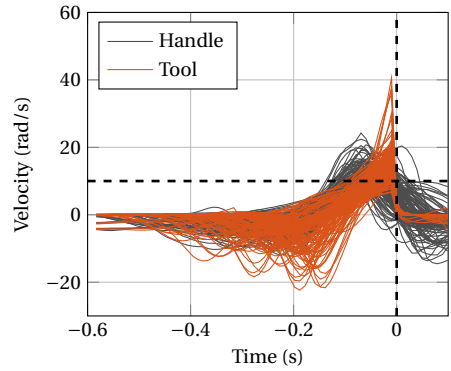}}
    \\[1cm]
    \subfloat[Adaptive]{\input{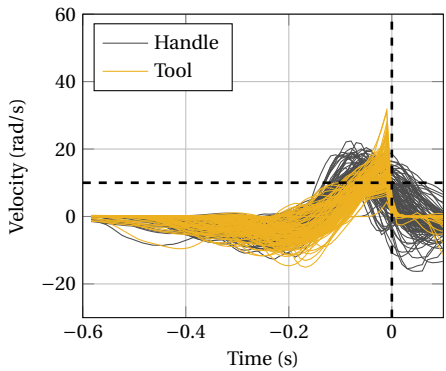}}
    \hspace{1cm}
    \subfloat[Adaptive*]{\input{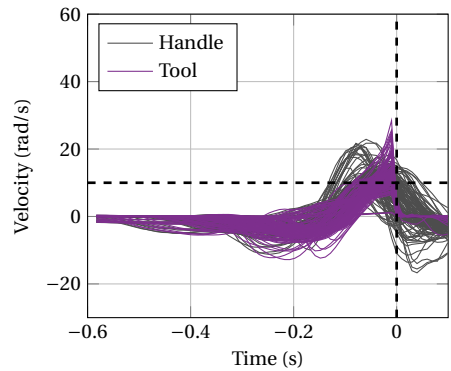}}
    \caption{Dynamic task handle and tool velocity plots of the reference measurements of all participants for trials in high, low and adaptive impedance setting, and of the task switch test for trials in the adaptive* setting. The dashed lines show the experiment criteria: the goal communicated to the participants was to achieve a maximum output velocity of at least \SI{10}{\radian\per\s}. The time axis was adjusted such that the time \SI{0}{\s} corresponded to the impact (the blue circle reaching the green box in Fig.~\ref{fig:variable-impedance-ui}.}
    \label{fig:variable-impedance-hammerplots}
\end{figure*}

\begin{figure}\RawFloats
\captionsetup[table]{position=top}
    \setlength{\figurewidth}{8cm}
    \setlength{\figureheight}{5cm}
    \footnotesize
    \centering
\colorlet{mycolor1}{matlab1}%
\colorlet{mycolor2}{matlab2}%
\colorlet{mycolor3}{matlab3}%
\colorlet{mycolor4}{matlab4}%
\begin{tikzpicture}

\begin{axis}[%
width=0.951\figurewidth,
height=\figureheight,
at={(0\figurewidth,0\figureheight)},
scale only axis,
xmin=0,
xmax=25,
xtick={ 1,  7, 13, 19, 24},
xlabel={Participant (-)},
ymin=0,
ymax=48,
ymajorgrids,
ylabel={Maximum Output Velocity (\si{\radian\per\s})},
legend style={legend cell align=left, align=left}
]
\draw[dashed, line width=0.1pt] (axis cs:0,10) -- (axis cs:25,10);
\addplot [color=mycolor1, draw=none, mark=*, mark options={solid, fill=mycolor1, mycolor1}]
 plot [error bars/.cd, y dir = both, y explicit]
 table[row sep=crcr, y error plus index=2, y error minus index=3]{%
1	15.7480931951375	0.809790736550312	0.929156502414358\\
2	15.0024449404237	1.293916157107	0.617798042914915\\
3	18.3581936459414	1.09852382698725	0.984498949828843\\
4	19.7275611853708	1.59008193379108	0.54203355579396\\
5	19.5115951733787	0.466835021583567	0.773055791139196\\
6	17.981019318058	0.860141926041898	0.969621551418982\\
7	12.3629134033465	0.618739047659242	0.688637621296158\\
8	10.1968521436431	1.21758023671725	1.66684879478885\\
9	10.6315424650605	0.933186098015248	0.527254127028751\\
10	15.1991342198081	0.833630590277355	1.10610215953348\\
11	20.9391689761418	0.895391842298725	1.360784976457\\
12	13.8481082281233	0.338595043668494	1.41204663529506\\
13	10.4028571704851	0.684059090751353	0.808329895762313\\
14	12.5194903905731	0.552956765960344	2.3575323066955\\
15	7.90361071222217	0.499092016066584	0.411208797078072\\
16	18.6484124411695	0.490198501821144	0.900726205267283\\
17	14.9260006024165	0.530966391043036	1.43575440057253\\
18	13.7611876170242	0.928808938280994	1.4401699768884\\
19	10.8228037162821	0.728960869359833	0.865707629797971\\
20	12.8445556696473	0.506971660978049	1.31554959485496\\
21	11.4398889760696	0.810575233405569	0.495930446588398\\
22	14.1705342412268	1.02613844496416	1.83806200291058\\
23	9.92340351112309	0.414699309791143	0.335158759485477\\
24	8.3865605774043	0.582571040924549	0.228593664874147\\
};
\addlegendentry{High Imp. Median}

\addplot [color=mycolor2, draw=none, mark=*, mark options={solid, fill=mycolor2, mycolor2}]
 plot [error bars/.cd, y dir = both, y explicit]
 table[row sep=crcr, y error plus index=2, y error minus index=3]{%
1	35.9249142187271	1.54413353104199	2.65914312769495\\
2	18.0954441962614	1.17090204883139	1.36392769449496\\
3	21.0762706602842	0.181840984457935	1.29833939573744\\
4	22.292002055388	1.0180417753317	1.71814580588951\\
5	14.8184163050972	1.06154136202542	1.12938169446035\\
6	16.0819280789426	2.25003482534668	0.950533588361068\\
7	14.2021943640569	1.25593712243889	0.903767819720409\\
8	15.5396838847339	0.43286739291835	1.08100398752124\\
9	18.5074287649357	2.75726818580279	4.08459642940927\\
10	13.584345309001	0.446399528738787	0.99361076257706\\
11	21.5411616061551	2.89674738303102	1.91311078217946\\
12	16.7466021712159	0.907741325280135	1.74817445740316\\
13	12.825786802072	0.937118027243528	0.774025140625447\\
14	16.444963049937	0.961048267264115	0.659567803722837\\
15	7.88675712360996	0.563541972425416	0.597479294633854\\
16	16.9182247882847	0.413384762158103	1.02258903059938\\
17	16.2516488868184	2.07196288429118	1.11773661904917\\
18	13.009305778998	1.27498746328467	1.05482170193858\\
19	21.359882792416	3.04326549314887	2.73606666006394\\
20	17.0575047777157	0.85712341617619	1.9847257364774\\
21	12.7609519313831	1.35402252625681	0.588967263770671\\
22	11.787231530213	1.24240642362881	0.729469592071233\\
23	12.0398511447361	0.581094914470158	1.12470371975663\\
24	8.31978176975388	1.12653341268151	0.750623070108915\\
};
\addlegendentry{Low Imp. Median}

\addplot [color=mycolor3, draw=none, mark=*, mark options={solid, fill=mycolor3, mycolor3}]
 plot [error bars/.cd, y dir = both, y explicit]
 table[row sep=crcr, y error plus index=2, y error minus index=3]{%
1	27.0209211197257	2.13230430818588	2.00509096278776\\
2	18.2875241357629	2.77402282822243	1.02202143400246\\
3	22.8080123601094	2.45637276094098	1.52044928586508\\
4	17.2327832410995	0.996269011770526	0.99416750040557\\
5	16.0958684940163	1.94570875521608	1.93300736544398\\
6	15.1272774509601	1.94843331287686	1.33899055923267\\
7	11.315602780137	0.614437995667597	1.56777727402103\\
8	10.9059270892533	1.19116398424742	0.851571051256318\\
9	17.1846830325422	1.7387196330765	1.98091570188873\\
10	13.5102148219015	1.67672504871626	2.17515415202307\\
11	25.0955890654161	1.7048302519414	3.07713149182654\\
12	14.9854562313815	2.06272831870952	1.47148137760703\\
13	13.8634941008052	0.647731487468944	1.58847189641608\\
14	15.9687905464271	0.524310765318361	2.56628228007089\\
15	8.14971234309693	0.870923845805633	0.521801365497975\\
16	16.8428866132387	1.35945619134189	2.15426148145743\\
17	15.9204208766597	3.74861843582031	1.83940733426273\\
18	10.570890549318	2.16904653122738	0.583248520360559\\
19	17.7419657992579	2.44411314878639	2.67326322885171\\
20	12.7482450000485	0.661017326183623	0.357056737008797\\
21	10.1889469566904	0.630285542395521	0.894378181180672\\
22	10.8993076673356	1.3719404704942	0.91559085627217\\
23	8.55747847533444	0.345597391020574	0.621929619984624\\
24	9.43179371855537	0.659091093686012	0.290834887691537\\
};
\addlegendentry{Adaptive Median}

\addplot [color=mycolor4, draw=none, mark=*, mark options={solid, fill=mycolor4, mycolor4}]
 plot [error bars/.cd, y dir = both, y explicit]
 table[row sep=crcr, y error plus index=2, y error minus index=3]{%
1	21.0350496230545	1.95103291709871	4.95072897155061\\
2	15.0865271582169	1.94733160039996	1.52755844751882\\
3	22.5985722365166	2.92585896229908	1.94026088201679\\
4	27.8206551243018	2.82344464168273	2.68237446634597\\
5	13.0028228573381	0.533842217038831	0.911178906003565\\
6	14.7173221859	3.10144478506228	1.31552179772535\\
7	10.9856519423943	0.785705528271212	1.32080112263251\\
8	13.0216190919505	2.73249420725596	0.947118520702885\\
9	14.3418631220838	1.67164065375275	1.11284098916684\\
10	10.3085140412509	0.897412129731848	0.332742989064256\\
11	25.8995715749922	3.85426175439105	7.69351827931403\\
12	15.4942410851378	1.50331388966234	0.918675541762909\\
13	11.0621624304909	2.72590198402899	2.11259515348191\\
14	13.4545593624823	3.5329818075292	1.24732827094833\\
15	8.70455215692412	0.859709077749466	0.512134480975984\\
16	13.8065921972945	1.93656624141922	1.39624413823817\\
17	15.290772717576	0.912341924226951	1.31338815633995\\
18	11.693184629298	0.670509748753281	0.907679785149556\\
19	14.6601304512582	1.84020761704054	1.54666529195521\\
20	13.7765088661294	0.855540992614733	1.99172067610282\\
21	11.3091265472332	0.540277305717995	0.774407188787841\\
22	12.5328876289146	0.762295733397352	1.05225951439895\\
23	8.86123400703975	1.05167078291695	0.96844406766072\\
24	7.53548768215982	0.878798848210506	0.989536563365695\\
};
\addlegendentry{Adaptive* Median}
\draw[<-,>=latex] (0.05\figurewidth,0.9\figureheight) -- (0.05\figurewidth,0.8\figureheight) node [midway,anchor=west] {Higher is better};
\end{axis}
\end{tikzpicture}%
    \caption{Per-participant median results of maximum output velocity measured during the dynamic task with \SI{95}{\percent} confidence intervals (higher is better).
    It was communicated to the participants that they should try to achieve a maximum output velocity of more than \SI{10}{\radian\per\s}.
    This is indicated by the dashed line.
    The expected outcome according to the experiment hypotheses was that the highest maximum output velocity would be achieved with the low impedance setting (red), the lowest with the low impedance (blue), and that the results with the adaptive (yellow) and adaptive* (purple) settings would not be significantly different from the results with the low impedance settings.
    }
    \label{fig:variable-impedance-participants-velocity-median}
\vspace{2cm}
    \captionof{table}{Grouped results over all participants of pairwise comparison of per-participant median maximum output velocities measured during the dynamic task, given as effect size of median difference and \SI{95}{\percent} confidence interval (positive difference indicates better performance of first setting compared to second, $n=24$, $p$-values refer to Wilcoxon's signed-rank test results).
    The last column indicates which experiment hypothesis was checked by the statistical test and whether the outcome supports (green) or contradicts (red) the hypothesis (significance level $\alpha = \SI{5}{\percent}$).
    }
    \label{tab:variable-impedance-results-velocity-median}
    \small
    % !TeX TS-program = lualatex
% !BIB TS-program = bibtex
% !TeX root = tab-results-velocity-median.tex
% !TeX encoding = UTF-8
% !TeX spellcheck = en_US
% Changelog:

\begin{tabular}{cL{0.11\columnwidth}cC{0.15\columnwidth}C{0.22\columnwidth}C{0.1\columnwidth}C{0.06\columnwidth}}
        \toprule
        \multicolumn{3}{c}{Impedance Setting Pair} & Median Diff. & \CI{} & $p$ & Hyp.\\
        \multicolumn{3}{c}{} & (\si{\radian\per\s}) & (\si{\radian\per\s}) & & \\
        \midrule
        \legendRectangle{LowImp} & Low -- High & \legendRectangle{HighImp} & 2.0 & [-0.017, 2.9] & 0.013 & \colorbox{colorHypConf}{B2} \\
        \midrule
       \legendRectangle{Adaptive} & Adaptive -- High & \legendRectangle{HighImp} & 0.48 & [-1.4, 3.3] & 0.34 & \colorbox{colorHypRej}{1.3} \\
       \legendRectangle{Adaptive} & Adaptive -- Low & \legendRectangle{LowImp} & -0.92 & [-2.6, -0.074] & 0.019 & \colorbox{colorHypRej}{1.4} \\
       \midrule
       \legendRectangle{Taskswitch} & Adaptive* -- High & \legendRectangle{HighImp} & 0.51  & [-1.1, 1.6]  & 0.55 & \colorbox{colorHypRej}{2.3} \\
       \legendRectangle{Taskswitch} & Adaptive* -- Low & \legendRectangle{LowImp} & -1.8 & [-3.1, -1.3] &  0.0059 & \colorbox{colorHypRej}{2.4} \\
        \bottomrule
    \end{tabular}
\end{figure}

\begin{figure}\RawFloats
\captionsetup[table]{position=top}
    \setlength{\figurewidth}{8cm}
    \setlength{\figureheight}{5cm}
    \footnotesize
    \centering
\colorlet{mycolor1}{matlab1}%
\colorlet{mycolor2}{matlab2}%
\colorlet{mycolor3}{matlab3}%
\colorlet{mycolor4}{matlab4}%
\begin{tikzpicture}

\begin{axis}[%
width=0.951\figurewidth,
height=\figureheight,
at={(0\figurewidth,0\figureheight)},
scale only axis,
xmin=0,
xmax=25,
xtick={ 1,  7, 13, 19, 24},
xlabel={Participant (-)},
ymin=0,
ymax=3.6,
ymajorgrids,
ylabel={Gain (-)},
legend style={legend cell align=left, align=left}
]
\draw[dashed, line width=0.1pt] (axis cs:0,1) -- (axis cs:25,1);
\addplot [color=mycolor1, draw=none, mark=*, mark options={solid, fill=mycolor1, mycolor1}]
 plot [error bars/.cd, y dir = both, y explicit]
 table[row sep=crcr, y error plus index=2, y error minus index=3]{%
1	1.01777867540542	0.0426994470694757	0.019208018349916\\
2	0.995622745365658	0.00998444999582548	0.00986536740375366\\
3	0.963102498515719	0.0234311484524254	0.028035465038647\\
4	1.0113748633268	0.0142162145710261	0.0114102040966343\\
5	1.01620106244616	0.0165279778268497	0.0380521014045739\\
6	0.963070005233598	0.031561432991886	0.0303067639761182\\
7	0.973340528501683	0.0268523669174967	0.00937173515490652\\
8	1.0490575812787	0.0193707252039264	0.0194574036029505\\
9	0.982001494569587	0.0100708086354442	0.0173219089392518\\
10	1.01506517318006	0.0114431014444181	0.0298523838059275\\
11	1.03291271498214	0.0217171198954038	0.0310948202614263\\
12	1.00817101015331	0.00733954565580053	0.0167004107736435\\
13	0.994895912003413	0.0160601134233571	0.00332061646745496\\
14	1.07957954713095	0.0138795668176444	0.00458276084732234\\
15	0.966580036905503	0.00586039457677934	0.0173975092755745\\
16	0.91645568067793	0.0457536806189106	0.051096612796097\\
17	0.990417834979795	0.0176364459007209	0.0234940731011847\\
18	0.968030384077772	0.0279441843433156	0.0237698157845098\\
19	1.01860433068432	0.00885154496179608	0.0205494685981268\\
20	0.91915761471095	0.00916712620871751	0.0252580988739605\\
21	0.976354866531892	0.0118429510149097	0.0230805797565277\\
22	0.910876161637472	0.0288440899611029	0.0589214572359354\\
23	0.967399889422681	0.00719748627766337	0.0145297403802724\\
24	0.921167413505853	0.0151467230475107	0.0554460596176305\\
};
\addlegendentry{High Imp. Median}

\addplot [color=mycolor2, draw=none, mark=*, mark options={solid, fill=mycolor2, mycolor2}]
 plot [error bars/.cd, y dir = both, y explicit]
 table[row sep=crcr, y error plus index=2, y error minus index=3]{%
1	2.56217370801905	0.166098684308292	0.353757047817362\\
2	1.26191928674442	0.120244456583748	0.123144321392178\\
3	0.957380075202724	0.0513922595707297	0.0716122002710067\\
4	1.03488854756898	0.0686899485727037	0.0833747467485202\\
5	0.88658779886774	0.0418943335273577	0.0935244168887911\\
6	0.968391427017521	0.0889835304724238	0.0672407366648403\\
7	1.25360417231516	0.0805945455797092	0.0534271980877359\\
8	1.78823668530632	0.0780878305472343	0.0895239393375284\\
9	1.94129130507188	0.268823797543935	0.287370431125556\\
10	0.95617114695973	0.0380587688368687	0.0576791666159784\\
11	1.18716358187846	0.108633705858295	0.173630141413623\\
12	1.35081975880476	0.11487396543499	0.137073943509079\\
13	1.29640636498018	0.0494862560286555	0.101858659084277\\
14	1.41904535318819	0.104764724543675	0.134718488610188\\
15	0.969969082232883	0.0392384201950677	0.0869345404682315\\
16	1.00033704485122	0.109307150684886	0.0652148186293702\\
17	1.07815570436136	0.0417962966656471	0.0735114818762934\\
18	0.940245991177513	0.0608928009253353	0.144135144113386\\
19	1.70256336904741	0.0568988017176351	0.245043987014422\\
20	1.0938015628641	0.144572302026368	0.083829738747629\\
21	1.03941746501497	0.112628232385294	0.0877524421468527\\
22	0.939771996713405	0.0574088531101187	0.0634758067626393\\
23	1.12308285995649	0.0866353485239313	0.0886176630434596\\
24	0.836970312895206	0.0858089262983033	0.0174176106360645\\
};
\addlegendentry{Low Imp. Median}

\addplot [color=mycolor3, draw=none, mark=*, mark options={solid, fill=mycolor3, mycolor3}]
 plot [error bars/.cd, y dir = both, y explicit]
 table[row sep=crcr, y error plus index=2, y error minus index=3]{%
1	1.73351371338722	0.102675094713784	0.0528985568719842\\
2	1.326934271362	0.0746796858110677	0.18793785322925\\
3	1.02905909435343	0.163538037404996	0.0775352039641566\\
4	0.905922505504507	0.0811596349127369	0.0512954723233463\\
5	0.80157510624761	0.0788160726753233	0.101079566966239\\
6	0.902592758980592	0.113368110341233	0.0849268520300103\\
7	0.868416716344721	0.0825771444596021	0.0367285507451398\\
8	1.3397755254167	0.0947691455718367	0.159241455412525\\
9	1.88659140013015	0.0792483992399737	0.275123831518127\\
10	0.993806542051493	0.100977232577587	0.0779441371706424\\
11	1.23611488400081	0.138265593812394	0.138328537739223\\
12	1.15084233265519	0.17262864894401	0.102663891687667\\
13	1.35528942336539	0.109970496383746	0.292273633064725\\
14	1.24948569355005	0.174569529603293	0.194463670730892\\
15	0.849432909339374	0.0724071182647257	0.0950188911515852\\
16	0.986502689737581	0.0535284571337398	0.107376535883098\\
17	1.13426878279041	0.144287796238292	0.271507815662751\\
18	0.725108233209312	0.112258523258462	0.06633466725139\\
19	1.47918512448007	0.128352448174195	0.211983511286286\\
20	0.800602960816149	0.066925187307485	0.0452323702437643\\
21	0.816181548831094	0.0642644974593181	0.0792780570792661\\
22	0.901889097293869	0.083407691078129	0.118885315054312\\
23	0.874431588705013	0.0581981150636171	0.0498979424037257\\
24	0.977319745128816	0.0329964831512055	0.0693577618120064\\
};
\addlegendentry{Adaptive Median}

\addplot [color=mycolor4, draw=none, mark=*, mark options={solid, fill=mycolor4, mycolor4}]
 plot [error bars/.cd, y dir = both, y explicit]
 table[row sep=crcr, y error plus index=2, y error minus index=3]{%
1	1.91730998613842	0.082543121933583	0.431037336313366\\
2	1.23254853829659	0.122765901994072	0.201593084392776\\
3	1.08309875391444	0.072794410177865	0.0686296279407341\\
4	1.33828997948791	0.106003841682304	0.196695878459237\\
5	0.759382802215761	0.0666352186769	0.0411852031059236\\
6	0.940264161170631	0.10143489443647	0.110390908150875\\
7	0.874568349838164	0.0564994727003503	0.126473575046816\\
8	1.5690066648349	0.188321846944636	0.131566120638752\\
9	1.86842649608714	0.093120439781472	0.163258693342418\\
10	0.904274097382042	0.0628897188058745	0.0494733044220624\\
11	1.27156319182718	0.140811793379215	0.391513241537551\\
12	1.38127679433596	0.155219805807116	0.166646332388052\\
13	1.12697107216671	0.268540670593723	0.131009201906118\\
14	1.1838124192688	0.295365472815389	0.121169401781408\\
15	0.912174035732409	0.0426052715903078	0.0852589486127312\\
16	0.848130572289954	0.169927058843958	0.0340621435510656\\
17	1.24531078582489	0.136731315359295	0.243716834404233\\
18	0.883284295997945	0.0463609974184215	0.10690170855421\\
19	1.3433520422949	0.213036728896803	0.21429881775769\\
20	0.824065354792205	0.0669849527816339	0.076827529878317\\
21	0.780821637080996	0.0373068066560875	0.0310909167686657\\
22	0.87337719538522	0.0596951356577664	0.0319074169421634\\
23	0.961713937949209	0.0945618343849124	0.0454696663795048\\
24	1.01315768272185	0.0296086526998662	0.0548553364893782\\
};
\addlegendentry{Adaptive* Median}
\draw[<-,>=latex] (0.05\figurewidth,0.9\figureheight) -- (0.05\figurewidth,0.8\figureheight) node [midway,anchor=west] {Higher is better};
\end{axis}
\end{tikzpicture}%
    \caption{Per-participant median results of gain measured during the dynamic task with \SI{95}{\percent} confidence intervals (higher is better).
    The dashed line indicates a gain of 1, which corresponds to the gain of a rigid actuator.
    The expected outcome according to the experiment hypotheses was that the highest gain would be achieved with the low impedance setting (red), the lowest with the low impedance (blue), and that the results with the adaptive (yellow) and adaptive* (purple) settings would not be significantly different from the results with the low impedance settings.
    }
    \label{fig:variable-impedance-participants-gain-median}
\vspace{2cm}
    \captionof{table}{Grouped results over all participants of pairwise comparison of per-participant median gains measured during the dynamic task, given as effect size of median difference and \SI{95}{\percent} confidence interval (positive difference indicates better performance of first setting compared to second, $n=24$, $p$-values refer to Wilcoxon's signed-rank test results).
    The last column indicates which experiment hypothesis was checked by the statistical test and whether the outcome supports (green) or contradicts (red) the hypothesis (significance level $\alpha = \SI{5}{\percent}$).
    }
    \label{tab:variable-impedance-results-gain-median}
    \small
    % !TeX TS-program = lualatex
% !BIB TS-program = bibtex
% !TeX root = tab-results-gain-median.tex
% !TeX encoding = UTF-8
% !TeX spellcheck = en_US
% Changelog:

\begin{tabular}{cL{0.11\columnwidth}cC{0.15\columnwidth}C{0.22\columnwidth}C{0.1\columnwidth}C{0.06\columnwidth}}
        \toprule
        \multicolumn{3}{c}{Impedance Setting Pair} & Median Diff. & \CI{} & $p$ & Hyp.\\
        \multicolumn{3}{c}{} & (-) & (-) & & \\
        \midrule
        \legendRectangle{LowImp} & Low -- High & \legendRectangle{HighImp} & 0.12 & [0.024, 0.28] & < 0.001 & \colorbox{colorHypConf}{B2} \\
        \midrule
        \legendRectangle{Adaptive} & Adaptive -- High & \legendRectangle{HighImp} & 0.061 & [-0.093, 0.17] & 0.17 & \colorbox{colorHypRej}{H1.3} \\
        \legendRectangle{Adaptive} & Adaptive -- Low & \legendRectangle{LowImp} & -0.10 & [-0.22, -0.014] & 0.0035 & \colorbox{colorHypRej}{H1.4}  \\
       \midrule
        \legendRectangle{Taskswitch} & Adaptive* -- High & \legendRectangle{HighImp} & 0.098  & [-0.054, 0.24]  & 0.049 & \colorbox{colorHypConf}{H2.3}  \\
        \legendRectangle{Taskswitch} & Adaptive* -- Low & \legendRectangle{LowImp} & -0.070 & [-0.17, -0.029] & 0.027 & \colorbox{colorHypRej}{H2.4}   \\
        \bottomrule
    \end{tabular}
\end{figure}

\begin{figure}\RawFloats
\captionsetup[table]{position=top}
    \setlength{\figurewidth}{8cm}
    \setlength{\figureheight}{5cm}
    \footnotesize
    \centering
\colorlet{mycolor1}{matlab1}%
\colorlet{mycolor2}{matlab2}%
\colorlet{mycolor3}{matlab3}%
\colorlet{mycolor4}{matlab4}%
\begin{tikzpicture}

\begin{axis}[%
width=0.951\figurewidth,
height=\figureheight,
at={(0\figurewidth,0\figureheight)},
scale only axis,
xmin=0,
xmax=25,
xtick={ 1,  7, 13, 19, 24},
xlabel={Participant (-)},
ymin=0,
ymax=57,
ymajorgrids,
ylabel={Maximum Output Velocity (\si{\radian\per\s})},
legend style={legend cell align=left, align=left}
]
\draw[dashed, line width=0.1pt] (axis cs:0,10) -- (axis cs:25,10);
\addplot [color=mycolor1, draw=none, mark=*, mark options={solid, fill=mycolor1, mycolor1}]
 plot [error bars/.cd, y dir = both, y explicit]
 table[row sep=crcr, y error plus index=2, y error minus index=3]{%
1	17.1740933629958	0.403932120391868	0.127736309208874\\
2	16.8363615927315	1.34391088262003	0.114588978666408\\
3	20.0448339848987	3.84767465267644	0.186968959991489\\
4	22.3027745764317	1.30951059304359	0.897222070812081\\
5	21.324997580855	0.184734541994985	0.567143871367591\\
6	20.4917464369637	3.33692243158794	1.27681185215207\\
7	14.062265946597	3.76199488758425	0.369771911129723\\
8	12.2541266974195	0.942427374380896	0.296056062414012\\
9	12.1284770264138	0.66346578547375	0.478233350478881\\
10	16.4008449779587	0.954535081571681	0.101416157347479\\
11	23.0019040031597	5.00592430095081	0.355094499492164\\
12	15.103738463028	1.12994644202541	0.379011784977926\\
13	12.0840791191374	0.260183266537096	0.455069927630291\\
14	14.3232398679641	0.558479338881172	0.686692226790603\\
15	10.0825455703376	3.568106498939	1.25106103249862\\
16	19.9702646027754	7.4315484055786	0.45641936274966\\
17	17.5153026679551	0.632398317257927	1.37934912251801\\
18	16.2712983618871	2.80957461144595	1.11811987983594\\
19	12.5679003665747	0.779912602439177	0.39156352434823\\
20	14.0958021499882	6.02791316713829	0.326943763292034\\
21	13.6484680010591	2.61415072328079	0.684422738814117\\
22	16.9667867895321	5.09442421568085	1.06443937908032\\
23	11.2462984841825	0.301046137622638	0.733865244464157\\
24	10.263167856904	3.34453335420564	0.550145572394927\\
};
\addlegendentry{High Imp. P90}

\addplot [color=mycolor2, draw=none, mark=*, mark options={solid, fill=mycolor2, mycolor2}]
 plot [error bars/.cd, y dir = both, y explicit]
 table[row sep=crcr, y error plus index=2, y error minus index=3]{%
1	41.3755980066656	4.05009249253855	1.8324521624909\\
2	20.4968156982074	2.0192482868416	0.427896384040707\\
3	23.2999250248716	0.934509565918496	1.47311780089025\\
4	26.1244382928796	0.501979394424136	2.12730706510174\\
5	18.0167393199652	2.74589803129813	0.565716594283245\\
6	19.7356742356562	2.27419724819503	0.989859167057329\\
7	17.320258753117	0.462704158253164	0.832814927222547\\
8	17.816613168938	1.21989689276827	0.456379319014946\\
9	25.3877391175272	5.35103136593159	1.19499425594236\\
10	15.3548028730968	1.59754559635301	0.706175964853808\\
11	25.8907184775736	0.536393729673676	1.27320175692277\\
12	18.4193835027983	3.3356819293511	0.731944585220077\\
13	14.7187876497532	1.05680935557348	0.733918551084106\\
14	19.043155442282	3.85009977483878	1.22361585285487\\
15	9.64974408521787	1.66483093907784	0.554368152174034\\
16	18.3326547299614	0.404599923949132	0.351906480611415\\
17	19.2195468147125	0.372398643170108	0.304284571403471\\
18	15.1117578640789	1.49908180599945	0.555912123651071\\
19	29.5388663827898	4.50900006839617	1.29388116776263\\
20	19.925407664539	2.33827433918182	0.720884600993461\\
21	14.8990980760129	0.448605070306181	0.385202236447331\\
22	13.6039491344295	0.202617280093182	0.521675481046783\\
23	14.113551111961	1.12090621824272	1.20412965860567\\
24	9.72686534098115	0.976797495202058	0.0393403671170347\\
};
\addlegendentry{Low Imp. P90}

\addplot [color=mycolor3, draw=none, mark=*, mark options={solid, fill=mycolor3, mycolor3}]
 plot [error bars/.cd, y dir = both, y explicit]
 table[row sep=crcr, y error plus index=2, y error minus index=3]{%
1	31.7753950783672	1.79598722447239	1.87174877500793\\
2	22.705919521088	0.58010975461632	0.353262557271357\\
3	27.4861296530728	4.07123764125118	0.896226640861496\\
4	20.0503088051545	4.59610090139956	1.22004882166203\\
5	19.3220195473816	2.23306972824936	0.330016202112393\\
6	18.2100775246745	1.69559711990803	0.728974015838677\\
7	13.0789330141534	1.44189798757714	0.275541189253897\\
8	14.6042321481553	1.6370389927343	0.26549340742465\\
9	20.2145759650669	3.96525356718566	1.02378645718541\\
10	19.1187708565656	1.47237588520902	2.65720982920886\\
11	31.2312711844878	4.51046450652846	3.41698820070805\\
12	19.3480527606645	0.961143558976072	0.957557181807022\\
13	15.9975503678077	0.63639299823943	0.363191148358085\\
14	20.0665540268648	2.55641376019381	1.33410101126546\\
15	9.86105573082682	1.26954975863831	0.170928687092946\\
16	19.7915194631916	2.84477770819474	1.08410762726167\\
17	20.9170079359814	0.362187530990401	0.541326109088168\\
18	13.5012421769654	1.57873998276258	0.281629390345598\\
19	21.6368559503165	2.24500532369373	0.481031155098993\\
20	15.0756025603668	1.85207570928522	0.198053384034827\\
21	11.4957083264002	0.314846494552564	0.438879141650073\\
22	13.0324417693484	0.705728059100741	0.38549671533508\\
23	10.4184638475141	0.435769834351172	0.50594242460177\\
24	11.1097664863076	0.639758545452164	0.467742160259467\\
};
\addlegendentry{Adaptive P90}

\addplot [color=mycolor4, draw=none, mark=*, mark options={solid, fill=mycolor4, mycolor4}]
 plot [error bars/.cd, y dir = both, y explicit]
 table[row sep=crcr, y error plus index=2, y error minus index=3]{%
1	29.4769899418272	1.30434359179005	2.23021854028254\\
2	19.3361680221496	2.04809531510896	1.08983083957465\\
3	27.2622550075704	0.803891937198124	1.19554956292014\\
4	34.4596780592376	2.68079957525016	0.419319721136915\\
5	14.1338036984493	2.0319281972469	0.147215736107169\\
6	18.863224147955	2.33616699098121	0.640898959958538\\
7	13.5597818272682	1.81013714309242	1.10608256809773\\
8	18.1956047063482	2.68225211410168	1.23995717902077\\
9	16.8890785249062	1.81846365213493	0.696014939300657\\
10	12.5107352364522	2.56022384596354	1.00590501996816\\
11	33.2277531563621	1.8238462931759	3.02431051250993\\
12	17.8903718545622	2.59674214834932	0.44016046722296\\
13	15.5150816118949	2.61318312050156	0.452859014786922\\
14	18.6860299667673	2.40920527101937	1.06268193822133\\
15	10.3558504852756	0.121055551408549	0.306554037660165\\
16	18.5175506298515	1.77384550490703	1.11844713386729\\
17	19.9221215580324	1.21845399674677	1.61923212390681\\
18	12.9822150231014	0.524203025533504	0.476698470048857\\
19	20.0645822273907	0.244145077860267	0.36586252256604\\
20	15.8821687450432	0.310879940440884	0.188986722701728\\
21	12.678707945963	1.25305838953135	0.0790780458675968\\
22	13.8534948626041	0.170476352301346	0.166440256370272\\
23	10.9599427728474	0.357297645018217	0.758368522775084\\
24	9.54528992753359	0.331196935103552	0.771337857102443\\
};
\addlegendentry{Adaptive* P90}
\draw[<-,>=latex] (0.05\figurewidth,0.9\figureheight) -- (0.05\figurewidth,0.8\figureheight) node [midway,anchor=west] {Higher is better};
\end{axis}
\end{tikzpicture}%
    \caption{Per-participant \percentile{90} of maximum output velocity measured during the dynamic task with \SI{95}{\percent} confidence intervals (higher is better).
    It was communicated to the participants that they should try to achieve a maximum output velocity of more than \SI{10}{\radian\per\s}.
    This is indicated by the dashed line.
    The expected outcome according to the experiment hypotheses was that the highest maximum output velocity would be achieved with the low impedance setting (red), the lowest with the low impedance (blue), and that the results with the adaptive (yellow) and adaptive* (purple) settings would not be significantly different from the results with the low impedance settings.
    }
    \label{fig:variable-impedance-participants-velocity-best}
\vspace{2cm}
    \captionof{table}{Grouped results over all participants of pairwise comparison of per-participant \percentile{90} maximum output velocities measured during the dynamic task, given as effect size of median difference and \SI{95}{\percent} confidence interval (positive difference indicates better performance of first setting compared to second, $n=24$, $p$-values refer to Wilcoxon's signed-rank test results).
    The last column indicates which experiment hypothesis was checked by the statistical test and whether the outcome supports (green) or contradicts (red) the hypothesis (significance level $\alpha = \SI{5}{\percent}$).
    }
    \label{tab:variable-impedance-results-velocity-best}
    \small
    % !TeX TS-program = lualatex
% !BIB TS-program = bibtex
% !TeX root = tab-results-velocity-best.tex
% !TeX encoding = UTF-8
% !TeX spellcheck = en_US
% Changelog:

\begin{tabular}{cL{0.11\columnwidth}cC{0.15\columnwidth}C{0.22\columnwidth}C{0.1\columnwidth}C{0.06\columnwidth}}
        \toprule
        \multicolumn{3}{c}{Impedance Setting Pair} & Median Diff. & \CI{} & $p$ & Hyp.\\
        \multicolumn{3}{c}{} & (\si{\radian\per\s}) & (\si{\radian\per\s}) & & \\
        \midrule
        \legendRectangle{LowImp} & Low -- High & \legendRectangle{HighImp} & 2.9 & [-0.43, 3.7] & 0.0039 & \colorbox{colorHypConf}{B2} \\
        \midrule
       \legendRectangle{Adaptive} & Adaptive -- High & \legendRectangle{HighImp} & 1.7 & [-0.83, 4.2] & 0.032 & \colorbox{colorHypConf}{H1.3} \\
       \legendRectangle{Adaptive} & Adaptive -- Low & \legendRectangle{LowImp} & -0.18 & [-3.4, 1.3] & 0.28 & \colorbox{colorHypConf}{H1.4} \\
       \midrule
       \legendRectangle{Taskswitch} & Adaptive* -- High & \legendRectangle{HighImp} & 2.1 & [-0.72, 4.4] & 0.069  & \colorbox{colorHypRej}{H2.3} \\
       \legendRectangle{Taskswitch} & Adaptive* -- Low & \legendRectangle{LowImp} & -0.70 & [-2.8, 0.25] & 0.11  & \colorbox{colorHypConf}{H2.4} \\
        \bottomrule
    \end{tabular}
\end{figure}

\begin{figure}\RawFloats
\captionsetup[table]{position=top}
    \setlength{\figurewidth}{8cm}
    \setlength{\figureheight}{5cm}
    \footnotesize
    \centering
\colorlet{mycolor1}{matlab1}%
\colorlet{mycolor2}{matlab2}%
\colorlet{mycolor3}{matlab3}%
\colorlet{mycolor4}{matlab4}%
\begin{tikzpicture}

\begin{axis}[%
width=0.951\figurewidth,
height=\figureheight,
at={(0\figurewidth,0\figureheight)},
scale only axis,
xmin=0,
xmax=25,
xtick={ 1,  7, 13, 19, 24},
xlabel={Participant (-)},
ymin=0,
ymax=4.8,
ymajorgrids,
ylabel={Gain (-)},
legend style={legend cell align=left, align=left}
]
\draw[dashed, line width=0.1pt] (axis cs:0,1) -- (axis cs:25,1);
\addplot [color=mycolor1, draw=none, mark=*, mark options={solid, fill=mycolor1, mycolor1}]
 plot [error bars/.cd, y dir = both, y explicit]
 table[row sep=crcr, y error plus index=2, y error minus index=3]{%
1	1.08929365413693	0.0291014551536644	0.00962095648927197\\
2	1.01808263807875	0.00905140007670324	0.00678363379936764\\
3	1.00812566841496	0.141342101090368	0.0158979233849758\\
4	1.04503521044862	0.0138695267365996	0.0114084974439665\\
5	1.04566715478041	0.0616402269713949	0.00634193121001014\\
6	1.01106827330246	0.288002459406738	0.00557633892998477\\
7	1.01114974917124	0.219321906048376	0.00645941113904702\\
8	1.1047231955359	0.0153362634412451	0.0166680894981732\\
9	1.00529691981153	0.0308575552701702	0.00704051308675047\\
10	1.04366526532549	0.0115723623253146	0.00382737438635172\\
11	1.06274831461837	0.301585701587428	0.00652034080516506\\
12	1.03152766922207	0.0265337305486359	0.00592481602455242\\
13	1.01844416535805	0.0523096970920265	0.00431672498699376\\
14	1.10350520207357	0.031105248463156	0.00287415218756393\\
15	0.986480749218942	0.262598525093952	0.00956040126786328\\
16	1.02521064957206	0.293089632616788	0.0265717510075117\\
17	1.02296514117072	0.0414345514894463	0.0063011216153257\\
18	1.023292329135	0.368021325170533	0.0085131574486117\\
19	1.04973042698262	0.00730664472706044	0.0184424492442432\\
20	0.951626523854683	0.378552958863779	0.014969490746103\\
21	1.01030885070221	0.298177713325136	0.0173884900647929\\
22	1.00958983565785	0.400715890702799	0.0174647743997133\\
23	0.984209136184567	0.0935430285172454	0.00577152400790026\\
24	0.94756796109785	0.24161941065955	0.00985831997225151\\
};
\addlegendentry{High Imp. P90}

\addplot [color=mycolor2, draw=none, mark=*, mark options={solid, fill=mycolor2, mycolor2}]
 plot [error bars/.cd, y dir = both, y explicit]
 table[row sep=crcr, y error plus index=2, y error minus index=3]{%
1	3.17998228488496	0.570871560487459	0.132748045080786\\
2	1.49921217659181	0.138128275875733	0.0676731403034989\\
3	1.02390177961784	0.126168581799229	0.00458320846887661\\
4	1.15929069869826	0.0466982651970691	0.0286749675677527\\
5	1.01588835198906	0.121345305748369	0.0776133051346483\\
6	1.19342731699608	0.17243350203724	0.121362729739609\\
7	1.42565613698402	0.149061822463129	0.0437817488958492\\
8	1.95045481750823	0.224703426496403	0.03358437658989\\
9	2.45855298308529	0.410398928704196	0.0358174461597716\\
10	1.06272375316472	0.0368467204251224	0.0274042586565382\\
11	1.37515119837898	0.0580337478712596	0.0704169381938977\\
12	1.61724774628606	0.248995611864012	0.102589709510573\\
13	1.55713378606312	0.0718930855420818	0.134787217399599\\
14	1.61676108100326	0.34712110673995	0.0623473356719466\\
15	1.1378071959869	0.0732718205350169	0.0328549556457152\\
16	1.17734509649196	0.0579545590936841	0.0537840436510535\\
17	1.25830181989476	0.0257665719471871	0.0457145678595867\\
18	1.12956615882583	0.0485075766407228	0.126195747397314\\
19	1.93645963846727	0.100592200492577	0.0331527465072357\\
20	1.34905604445325	0.183350779794994	0.0938482257920334\\
21	1.17626687226653	0.10729253065144	0.0167619313067868\\
22	1.09559920907436	0.0739527139611333	0.0446483970313847\\
23	1.26529003958843	0.0320481758790412	0.027438067884562\\
24	1.05995235945403	0.176444337794853	0.108486345099075\\
};
\addlegendentry{Low Imp. P90}

\addplot [color=mycolor3, draw=none, mark=*, mark options={solid, fill=mycolor3, mycolor3}]
 plot [error bars/.cd, y dir = both, y explicit]
 table[row sep=crcr, y error plus index=2, y error minus index=3]{%
1	1.96290178682526	0.140678654662782	0.0467957421112017\\
2	1.47441527128494	0.0918369684219353	0.0398615620739406\\
3	1.30233731801699	0.0055464656435642	0.0257415133608607\\
4	1.18417197760078	0.159479641915408	0.154282019811513\\
5	0.965458921789902	0.0978647211235336	0.044018283349529\\
6	1.22893607447701	0.217490929800182	0.137338126103878\\
7	1.00421461979534	0.0478169232222438	0.0438043461794718\\
8	1.556932121571	0.0639981485730596	0.0554768417582943\\
9	2.12254172695855	0.167545301175749	0.0882795187394567\\
10	1.1723883257231	0.196443841346438	0.00209259350420954\\
11	1.5418574475744	0.275079007306602	0.109186506540858\\
12	1.43551909487679	0.160040660874997	0.0533693383116698\\
13	1.57194313885085	0.0495183206732464	0.0209793277310399\\
14	1.50306501552404	0.16939595751302	0.0433780260251582\\
15	0.996507560876749	0.0592409272838998	0.0473857480295603\\
16	1.13801875825941	0.153419928639104	0.0307226003299614\\
17	1.43330568565496	0.0714295622943348	0.0710346943631466\\
18	0.902009621313727	0.0462873122905499	0.0437992544999397\\
19	1.75164295389171	0.126534706866253	0.0724579731219519\\
20	0.913423151927697	0.0406327891960062	0.0140894366053858\\
21	0.943554722010789	0.00999685410078666	0.0257574061942231\\
22	1.04838174383394	0.0077919961813615	0.0353231762995079\\
23	1.03552328054801	0.028617276497604	0.0660845024075415\\
24	1.05548820632219	0.0820868313308996	0.0126814116971186\\
};
\addlegendentry{Adaptive P90}

\addplot [color=mycolor4, draw=none, mark=*, mark options={solid, fill=mycolor4, mycolor4}]
 plot [error bars/.cd, y dir = both, y explicit]
 table[row sep=crcr, y error plus index=2, y error minus index=3]{%
1	2.10741692507757	0.525570045164456	0.0365439629623401\\
2	1.38611924272188	0.233900707070434	0.0232001682896765\\
3	1.27702849983129	0.149469777855871	0.0377230850163852\\
4	1.58246794964739	0.145679085051296	0.069850522524004\\
5	0.894772131027928	0.0377311794238532	0.0456981568286591\\
6	1.13674777854847	0.146496756397373	0.0430961158842118\\
7	1.00950015466978	0.054127342963654	0.0607397244006075\\
8	1.93338967757354	0.0423944986591729	0.110169001339916\\
9	2.10021835824066	0.00140602498420073	0.116085097039331\\
10	1.08469728142806	0.305246243959424	0.0590482348158816\\
11	1.53551139573082	0.160647681457442	0.0776314423929336\\
12	1.61212546947217	0.0435569918636178	0.0413372346074423\\
13	1.61178662733138	0.156525139306324	0.0566658782287417\\
14	1.6024386420604	0.160280647522572	0.0396883880437657\\
15	1.02990585912933	0.0394104302554732	0.0321033749689346\\
16	1.10339675020357	0.0888622811616691	0.0598161718797403\\
17	1.48504455717426	0.0131652479165476	0.0920686096791683\\
18	1.00200355739011	0.0209971594699205	0.0632693265499464\\
19	1.74254752876556	0.0650012424095801	0.0650409845590814\\
20	0.965532118566187	0.0209895594006434	0.0338767630690989\\
21	0.877880012709994	0.0754177815076053	0.0331717021817256\\
22	1.03065792863374	0.0173858048818358	0.0230920328413193\\
23	1.13238658266699	0.0251883120872485	0.0291785999920551\\
24	1.06615646367079	0.0435315843958495	0.0150680430189598\\
};
\addlegendentry{Adaptive* P90}
\draw[<-,>=latex] (0.05\figurewidth,0.9\figureheight) -- (0.05\figurewidth,0.8\figureheight) node [midway,anchor=west] {Higher is better};
\end{axis}
\end{tikzpicture}%
    \caption{Per-participant \percentile{90} of gain measured during the dynamic task with \SI{95}{\percent} confidence intervals (higher is better).
    The dashed line indicates a gain of 1, which corresponds to the gain of a rigid actuator.
    The expected outcome according to the experiment hypotheses was that the highest gain would be achieved with the low impedance setting (red), the lowest with the low impedance (blue), and that the results with the adaptive (yellow) and adaptive* (purple) settings would not be significantly different from the results with the low impedance settings.
    }
    \label{fig:variable-impedance-participants-gain-best}
\vspace{2cm}
    \captionof{table}{Grouped results over all participants of pairwise comparison of per-participant  \percentile{90} gains measured during the dynamic task, given as effect size of median difference and \SI{95}{\percent} confidence interval (positive difference indicates better performance of first setting compared to second, $n=24$, $p$-values refer to Wilcoxon's signed-rank test results).
    The last column indicates which experiment hypothesis was checked by the statistical test and whether the outcome supports (green) or contradicts (red) the hypothesis (significance level $\alpha = \SI{5}{\percent}$).
    }
    \label{tab:variable-impedance-results-gain-best}
    \small
    % !TeX TS-program = lualatex
% !BIB TS-program = bibtex
% !TeX root = tab-results-gain-best.tex
% !TeX encoding = UTF-8
% !TeX spellcheck = en_US
% Changelog:

\begin{tabular}{cL{0.11\columnwidth}cC{0.15\columnwidth}C{0.22\columnwidth}C{0.1\columnwidth}C{0.06\columnwidth}}
        \toprule
        \multicolumn{3}{c}{Impedance Setting Pair} & Median Diff. & \CI{} & $p$ & Hyp.\\
        \multicolumn{3}{c}{} & (-) & (-) & & \\
        \midrule
        \legendRectangle{LowImp} & Low -- High & \legendRectangle{HighImp} & 0.26 & [0.15, 0.48] & < 0.001 & \colorbox{colorHypConf}{B2} \\
        \midrule
        \legendRectangle{Adaptive} & Adaptive -- High & \legendRectangle{HighImp} & 0.18 & [0.051, 0.41] & < 0.001 & \colorbox{colorHypConf}{H1.3} \\
        \legendRectangle{Adaptive} & Adaptive -- Low & \legendRectangle{LowImp} & -0.082 & [-0.23, -0.0045] & 0.013 & \colorbox{colorHypRej}{H1.4} \\
        \midrule
        \legendRectangle{Taskswitch} & Adaptive* -- High & \legendRectangle{HighImp} & 0.21 & [0.043, 0.50] & < 0.001 & \colorbox{colorHypConf}{H2.3} \\
        \legendRectangle{Taskswitch} & Adaptive* -- Low & \legendRectangle{LowImp} & -0.069 & [-0.13, -0.0051] & 0.069 & \colorbox{colorHypConf}{H2.4} \\
        \bottomrule
    \end{tabular}
\end{figure}

Fig.~\ref{fig:variable-impedance-hammerplots} shows the plots of the handle and tool velocity during the reference measurements and the task switch test of all participants. It gives a qualitative comparison between the three settings of the reference measurement plus the task switch test (adaptive*).
As can be seen in the plots for the low impedance and adaptive settings, the tool velocity on impact at time zero is often considerably higher than the handle velocity, whereas it is very close to the handle velocity for the stiff setting.

Fig.~\ref{fig:variable-impedance-participants-velocity-median} to \ref{fig:variable-impedance-participants-gain-best} show the per-participant median and \percentiles{90} results of maximum output velocities and gains with \SI{95}{\percent} confidence intervals (higher is better).
The expected outcome according to the experiment hypotheses was that the highest maximum output velocity and gain would be achieved with the low impedance setting (red), the lowest with the high impedance (blue), and that the results with the adaptive (yellow) and adaptive* (purple) settings would not be significantly different from the results with the low impedance settings.

The grouped statistical comparison over all participants of maximum output velocity medians (cf. table \ref{tab:variable-impedance-results-velocity-median}) showed a significant difference at the \siglevel{5} between low and high impedance settings (supporting hypothesis B2).
It showed no significant difference at the \siglevel{20} between adaptive and high impedance settings, and adaptive* and high impedance settings, but a significant difference at the \siglevel{5} between adaptive and low impedance settings and at the \siglevel{1} between adaptive* and low impedance settings (contradicting hypotheses H1.3, H1.4, H2.3 and H2.4).

The grouped statistical comparison over all participants of gain medians (cf. table \ref{tab:variable-impedance-results-gain-median}) showed a significant difference at the \siglevel{0.1} between low and high impedance settings (supporting hypothesis B2).
It showed no significant difference at the \siglevel{10} between adaptive and high impedance settings, but a significant difference at the \siglevel{5} between  adaptive* and high impedance settings, at the \siglevel{0.5} between adaptive and low impedance settings, and at the \siglevel{5} between adaptive* and low impedance settings (supporting hypothesis H2.3, but contradicting hypotheses H1.3, H1.4, and H2.4).

The grouped statistical comparison over all participants of maximum output velocity 90$^{th}$ percentiles (cf. table \ref{tab:variable-impedance-results-velocity-best}) showed a significant difference at the \siglevel{0.5} between low and high impedance settings (supporting hypothesis B2).
It showed a significant difference at the \siglevel{5} between adaptive and high impedance setting, but no significant difference at the \siglevel{5} between adaptive and low impedance settings, at the \siglevel{5} between adaptive* and high impedance settings, and at the \siglevel{10} between adaptive* and low impedance settings (supporting hypotheses H1.3, H1.4 and H2.4, but contradicting hypothesis H2.3).

The grouped statistical comparison over all participants of gain 90$^{th}$ percentiles (cf. table \ref{tab:variable-impedance-results-gain-best}) showed a significant difference at the \siglevel{0.1} between low and high impedance settings (supporting hypothesis B2).
It showed a significant difference at the \SI{0.1}{\percent}-level between adaptive and high impedance settings, and at the \siglevel{0.1} between adaptive* and high impedance settings, a significant difference at the \siglevel{5} between adaptive and low impedance settings, but no significant difference at the \siglevel{5} between adaptive* and low impedance settings (supporting hypotheses H1.3, H2.3 and H2.4, but contradicting hypothesis H1.4).

\section{Discussion}
\label{sec:variable-impedance-discussion}
This study shows the potential and the limits of the proposed self-adapting impedance variation algorithm for telemanipulation of variable impedance tool devices.
It is also the first human-user study with the new Dyrac \gls{via} \cite{DyracPaper}, although in a simulation of the actuator, evaluating a possibility to use one actuator for precision and dynamic tasks instead of using a dedicated actuator for each type of tasks.

Using the median results as reference for medium performance, the study shows that normally high impedance tool devices are more appropriate for free air precision tasks and low impedance tool devices are more appropriate for dynamic tasks (hypotheses B1 and B2).
This is not surprising, as the first is well established state of the art in robot design, and we showed the second in two previous experimental studies with \glspl{sea} \cite{Aiple2018}.
The results also strongly indicate that the experiment participants were able to achieve the same medium performance in the precision task with the new adaptive impedance as with the conventional high impedance.
This is an encouraging result as precision tasks play a very important role in teleoperation (\eg for pick-and-place tasks, inspection, insertion, etc.), such that it can be argued that a tool device not performing well in precision tasks would be very impractical.

On the other hand, it is surprising that the human users seem to have been able to achieve similar performance on the precision task with a low impedance tool device as with a high impedance tool device in their very good trials (\percentile{10} results).
Apparently, they were able in the very good trials to sufficiently suppress the oscillations of the low impedance tool device to come to rest at the target position in a comparable time to the very good trials with the high impedance tool device.
The observation that they only achieved this in their very good trials suggests that reaching better performance required more effort with low tool device impedance than with high tool device impedance.
However, this doubt on the basic assumption (hypothesis B1) for the precision task in the case of very good trials makes it difficult to conclude about the performance with the adaptive setting on the precision task for very good trials.

This might also have been influenced by the bounded by zero nature of the metrics, that gave a lower limit to the travel time and integral of time-weighted absolute error (ITAE) even under very good conditions (\percentile{10} of high impedance setting, cf. figure~\ref{fig:variable-impedance-participants-traveltime-best}, table~\ref{tab:variable-impedance-results-traveltime-best}, figure~\ref{fig:variable-impedance-participants-itae-best} and table \ref{tab:variable-impedance-results-itae-best}).
This could have facilitated that the participants achieved similar performance for their very good trials of less good conditions.
The particularly good performance on the precision task during the task switch phase of the experiment could be due to the fact that the participants had performed more trials at this point of the experiment or that they performed less trials in a row.
However, no concluding explanation was found on why they reached even better performance than the high impedance performance of the reference measurement phase.

Although a significantly better medium performance in terms of maximum output velocity and gain was measured on the dynamic task with low tool device impedance than with high tool impedance, the observed effect sizes were much smaller than in the experiments with \glspl{sea} \cite{Aiple2018}.
The effect size on the maximum output velocity in \ExpOne of \cite{Aiple2018} was \SI{14.34}{\radian\per\s}, \CI $= [11.30, 15.43]$ \si{\radian\per\s} at a resonance frequency of \SI{4.8}{\Hz} and \SI{9.88}{\radian\per\s}, \CI = $[8.00 , 16.72]$ \si{\radian\per\s} at \SI{6.9}{\Hz}.
The effect size on the gain was \SI{1.16}{}, \CI $= [0.93, 1.36]$ at \SI{4.8}{\Hz} and \SI{1.31}{}, \CI $= [0.73, 1.51]$ at \SI{6.9}{\Hz}.
The effect size on the maximum output velocity in the medium performance measurement of this study was only \SI{2}{\radian\per\s}, \CI $= [-0.017, 2.9]$ \si{\radian\per\s} at a resonance frequency of \SI{4.5}{\Hz} (low impedance), and \SI{0.48}{\radian\per\s}, \CI $= [-1.4, 3.3]$ \si{\radian\per\s} at \SI{6.5}{\Hz} (no significant effect for the adaptive impedance).
The effect size on the gain in this study was only \SI{0.12}{}, \CI $= [0.024, 0.28]$ at \SI{4.5}{\Hz} (low impedance), and \SI{0.061}{}, \CI $= [-0.093, 0.17]$ \si{} at \SI{6.5}{\Hz} (no significant effect for the adaptive impedance).

Looking at figure~\ref{fig:variable-impedance-participants-velocity-median} and \ref{fig:variable-impedance-participants-gain-median}, it stands out that some participants obtained a median gain of less than 1 with the low impedance setting (\eg participant 3, 5, 18, 22, 24), which means that did not manage to exploit the resonance effect to achieve higher performance.
Whereas others achieved gains of 2 and more (\eg participant 1, 8, 9, 19), which is closer to our previous results.
This suggests that it was not obvious to the participants how they could use the elasticity of the tool to their advantage, even if some participants still performed visibly better in their very good trials with the adaptive settings than for the median (\eg participant 12, 13, 14).
This might be due to the different instructions, as in the previous study the participants saw an instruction video explaining the principles of resonance extensively \cite{Aiple2017a}.
That could be an indication that the elastic hammering task is less intuitive than we concluded previously, requiring some understanding of the resonance effect even if no extensive training.

It could also be an effect of the different experiment apparatus and of the impedance variation algorithm.
Especially for the adaptive setting the fact that the actuator impedance is still high at the beginning of the motion and only becomes lower at higher velocities might make it difficult to anticipate the behavior of the actuator for dynamic tasks.
Also the system identification of the adaptive setting showed a resonance frequency close to \SI{6.5}{\Hz} instead of the \SI{4.5}{\Hz} resonance frequency of the low impedance setting.
This might also have degraded the performance with the adaptive setting compared to the low impedance setting, as a different, faster motion was required to optimally excite the resonance of the actuator under the adaptive setting than under the low impedance setting.
Indeed, in \ExpOne of \cite{Aiple2018} the confidence intervals of the effect sizes were larger at a resonance frequency of \SI{6.9}{\Hz}, which supports this hypothesis.
This seems coherent with the indications of this study that the experiment participants were only able to achieve significantly better performance in some cases with the adaptive setting compared to the high impedance setting.
However, the effect sizes are also small for those measures that show a significant effect (cf. table~\ref{tab:variable-impedance-results-velocity-best} and \ref{tab:variable-impedance-results-gain-best}), making it impossible to formulate strong statements of any performance improvements with the adaptive setting.

Overall, the experiment results could only partially support the experiment hypotheses, namely the median results of the precision task strongly indicate that the participants could achieve equal performance with the adaptive setting as with the high impedance setting.
Even if the adaptive setting did not yield consistent results for the dynamic task, it also has the advantage that it has a low impedance on high speed impacts, thereby increasing safety in human-robot cooperation \cite{Bicchi2004} and helping to reduce wear on the actuator \cite{Pratt1997}.
From a system point of view these can be considered positive results for the adaptive impedance variation principle for dynamic teleoperation compared to the state of the art rigid tool devices.

\section{Conclusion}
This study demonstrated the proof of principle for self-adapting impedance control without additional sensors measuring the operator's limb stiffness, where the operator does not need to consciously change from one mode of operation to another to perform different kinds of tasks.
A self-adapting impedance variation control law was presented, and its use demonstrated by a human-subject experiment.
This control law automatically changes the impedance of a variable impedance actuator tool device in a teleoperation system depending on the velocity of the handle device.
The experiment validated this control law for the use of one \gls{via} tool device for precision tasks and dynamic tasks without the need of additional sensors for measuring the intended stiffness, reducing system complexity.
The performance obtained for precision task was comparable with the performance achieved with a high impedance actuator.
The performance obtained for dynamic tasks was not as good as observed in previous experiments with a permanently soft actuator.

Hence, the system can be used equally well as a rigid system for precision tasks, which constitute the classic domain of teleoperation.
Additionally, the system is better suited for dynamic tasks than rigid teleoperation systems, as the safety is increased in human-robot cooperation scenarios and the wear is reduced of the \gls{via} by automatically decreasing the impedance when operating at higher speed.
These are promising results for the development of wide-range teleoperation systems, able to cover the full dynamic spectrum of the human motion bandwidth.

\section{Acknowledgements}
\label{sec:acknowledgements}
This project was supported by the Dutch organization for scientific research (NWO) under the project grant~12161.


\begin{thebibliography}{10}
\expandafter\ifx\csname url\endcsname\relax
  \def\url#1{\texttt{#1}}\fi
\expandafter\ifx\csname urlprefix\endcsname\relax\def\urlprefix{URL }\fi
\expandafter\ifx\csname href\endcsname\relax
  \def\href#1#2{#2} \def\path#1{#1}\fi

\bibitem{Vanderborght2012}
B.~Vanderborght, A.~Bicchi, E.~Burdet, D.~Caldwell, R.~Carloni, M.~Catalano,
  G.~Ganesh, M.~Garabini, M.~Grebenstein, G.~Grioli, S.~Haddadin, A.~Jafari,
  M.~Laffranchi, D.~Lefeber, F.~Petit, S.~Stramigioli, N.~Tsagarakis, M.~V.
  Damme, R.~V. Ham, L.~C. Visser, S.~Wolf, A.~Albu-Sch{\"{a}}ffer,
  \href{http://ieeexplore.ieee.org/ielx5/6363628/6385431/06385433.pdf?tp=&arnumber=6385433&isnumber=6385431}{{Variable
  impedance actuators: Moving the robots of tomorrow}}, in: Intelligent Robots
  and Systems (IROS), 2012 IEEE/RSJ International Conference on, 2012, pp.
  5454--5455 (2012).
\newblock \href {https://doi.org/10.1109/IROS.2012.6385433}
  {\path{doi:10.1109/IROS.2012.6385433}}.
\newline\urlprefix\url{http://ieeexplore.ieee.org/ielx5/6363628/6385431/06385433.pdf?tp=&arnumber=6385433&isnumber=6385431}

\bibitem{Vanderborght2013}
B.~Vanderborght, A.~Albu-Sch{\"{a}}ffer, A.~Bicchi, E.~Burdet, D.~G. Caldwell,
  R.~Carloni, M.~Catalano, O.~Eiberger, W.~Friedl, G.~Ganesh, M.~Garabini,
  M.~Grebenstein, G.~Grioli, S.~Haddadin, H.~Hoppner, A.~Jafari, M.~Laffranchi,
  D.~Lefeber, F.~Petit, S.~Stramigioli, N.~Tsagarakis, M.~Van~Damme,
  R.~Van~Ham, L.~C. Visser, S.~Wolf,
  \href{http://www.sciencedirect.com/science/article/pii/S0921889013001188}{{Variable
  impedance actuators: A review}}, Robotics and Autonomous Systems 61~(12)
  (2013) 1601--1614 (2013).
\newblock \href {https://doi.org/10.1016/j.robot.2013.06.009}
  {\path{doi:10.1016/j.robot.2013.06.009}}.
\newline\urlprefix\url{http://www.sciencedirect.com/science/article/pii/S0921889013001188}

\bibitem{Wolf2016}
S.~Wolf, G.~Grioli, O.~Eiberger, W.~Friedl, M.~Grebenstein, H.~Hoppner,
  E.~Burdet, D.~G. Caldwell, R.~Carloni, M.~G. Catalano, D.~Lefeber,
  S.~Stramigioli, N.~Tsagarakis, M.~Van~Damme, R.~Van~Ham, B.~Vanderborght,
  L.~C. Visser, A.~Bicchi, A.~Albu-Schaffer,
  \href{http://ieeexplore.ieee.org/document/7330025/}{{Variable Stiffness
  Actuators: Review on Design and Components}}, IEEE/ASME Transactions on
  Mechatronics 21~(5) (2016) 2418--2430 (10 2016).
\newblock \href {https://doi.org/10.1109/TMECH.2015.2501019}
  {\path{doi:10.1109/TMECH.2015.2501019}}.
\newline\urlprefix\url{http://ieeexplore.ieee.org/document/7330025/}

\bibitem{Hogan1984}
N.~Hogan, \href{http://ieeexplore.ieee.org/document/1103644/}{{Adaptive control
  of mechanical impedance by coactivation of antagonist muscles}}, IEEE
  Transactions on Automatic Control 29~(8) (1984) 681--690 (8 1984).
\newblock \href {https://doi.org/10.1109/TAC.1984.1103644}
  {\path{doi:10.1109/TAC.1984.1103644}}.
\newline\urlprefix\url{http://ieeexplore.ieee.org/document/1103644/}

\bibitem{DyracPaper}
M.~Aiple, W.~Gregoor, A.~Schiele, {A Dynamic Robotic Actuator with Variable
  Physical Stiffness and Damping}, Mechanism and Machine Theory (submitted for
  publication) (2020).

\bibitem{Semini2015}
C.~Semini, V.~Barasuol, T.~Boaventura, M.~Frigerio, M.~Focchi, D.~G. Caldwell,
  J.~Buchli,
  \href{http://journals.sagepub.com/doi/10.1177/0278364915578839}{{Towards
  versatile legged robots through active impedance control}}, The International
  Journal of Robotics Research 34~(7) (2015) 1003--1020 (6 2015).
\newblock \href {https://doi.org/10.1177/0278364915578839}
  {\path{doi:10.1177/0278364915578839}}.
\newline\urlprefix\url{http://journals.sagepub.com/doi/10.1177/0278364915578839}

\bibitem{Buchli2011}
J.~Buchli, E.~Theodorou, F.~Stulp, S.~Schaal,
  \href{http://citeseerx.ist.psu.edu/viewdoc/download?doi=10.1.1.167.7254&rep=rep1&type=pdf}{{Variable
  impedance control a reinforcement learning approach}}, Robotics: Science and
  Systems VI (2011) 153--160 (2011).
\newline\urlprefix\url{http://citeseerx.ist.psu.edu/viewdoc/download?doi=10.1.1.167.7254&rep=rep1&type=pdf}

\bibitem{Braun2012}
D.~Braun, M.~Howard, S.~Vijayakumar,
  \href{http://dx.doi.org/10.1007/s10514-012-9302-3
  http://link.springer.com/10.1007/s10514-012-9302-3}{{Optimal variable
  stiffness control: formulation and application to explosive movement tasks}},
  Autonomous Robots 33~(3) (2012) 237--253 (10 2012).
\newblock \href {https://doi.org/10.1007/s10514-012-9302-3}
  {\path{doi:10.1007/s10514-012-9302-3}}.
\newline\urlprefix\url{http://dx.doi.org/10.1007/s10514-012-9302-3
  http://link.springer.com/10.1007/s10514-012-9302-3}

\bibitem{VanTeeffelen2018}
K.~Van~Teeffelen, D.~Dresscher, W.~Van~Dijk, S.~Stramigioli, {Intuitive
  Impedance Modulation in Haptic Control Using Electromyography}, in: 2018 7th
  IEEE International Conference on Biomedical Robotics and Biomechatronics
  (Biorob), IEEE, 2018, pp. 1211--1217 (2018).

\bibitem{Hill2009}
M.~D. Hill, G.~Niemeyer, {Real-time estimation of human impedance for haptic
  interfaces}, in: EuroHaptics conference, 2009 and Symposium on Haptic
  Interfaces for Virtual Environment and Teleoperator Systems. World Haptics
  2009. Third Joint, 2009, pp. 440--445 (2009).
\newblock \href {https://doi.org/10.1109/WHC.2009.4810893}
  {\path{doi:10.1109/WHC.2009.4810893}}.

\bibitem{Walker2010a}
D.~S. Walker, R.~P. Wilson, G.~Niemeyer,
  \href{http://ieeexplore.ieee.org/ielx5/5501116/5509124/05509811.pdf?tp=&arnumber=5509811&isnumber=5509124}{{User-controlled
  variable impedance teleoperation}}, in: Robotics and Automation (ICRA), 2010
  IEEE International Conference on, 2010, pp. 5352--5357 (2010).
\newblock \href {https://doi.org/10.1109/ROBOT.2010.5509811}
  {\path{doi:10.1109/ROBOT.2010.5509811}}.
\newline\urlprefix\url{http://ieeexplore.ieee.org/ielx5/5501116/5509124/05509811.pdf?tp=&arnumber=5509811&isnumber=5509124}

\bibitem{Bicchi2004}
A.~Bicchi, G.~Tonietti,
  \href{http://ieeexplore.ieee.org/document/1310939/}{{Fast and "Soft-Arm"
  Tactics}}, IEEE Robotics {\&} Automation Magazine 11~(2) (2004) 22--33 (6
  2004).
\newblock \href {https://doi.org/10.1109/MRA.2004.1310939}
  {\path{doi:10.1109/MRA.2004.1310939}}.
\newline\urlprefix\url{http://ieeexplore.ieee.org/document/1310939/}

\bibitem{Aiple2018}
M.~Aiple, J.~Smisek, A.~Schiele,
  \href{https://ieeexplore.ieee.org/document/8540940/}{{Increasing Impact by
  Mechanical Resonance for Teleoperated Hammering}}, IEEE Transactions on
  Haptics (2018) 1--1 (2018).
\newblock \href {https://doi.org/10.1109/TOH.2018.2882401}
  {\path{doi:10.1109/TOH.2018.2882401}}.
\newline\urlprefix\url{https://ieeexplore.ieee.org/document/8540940/}

\bibitem{Aiple2017}
M.~Aiple, A.~Schiele, \href{http://ieeexplore.ieee.org/document/7989896/
  http://ieeexplore.ieee.org/document/7989896/?reload=true
  http://arxiv.org/abs/1703.09934 http://www.arxiv.org/pdf/1703.09934.pdf
  https://arxiv.org/abs/1703.09934}{{Towards teleoperation with human-like
  dynamics: Human use of elastic tools}}, in: 2017 IEEE World Haptics
  Conference (WHC), IEEE, 2017, pp. 171--176 (6 2017).
\newblock \href {https://doi.org/10.1109/WHC.2017.7989896}
  {\path{doi:10.1109/WHC.2017.7989896}}.
\newline\urlprefix\url{http://ieeexplore.ieee.org/document/7989896/
  http://ieeexplore.ieee.org/document/7989896/?reload=true
  http://arxiv.org/abs/1703.09934 http://www.arxiv.org/pdf/1703.09934.pdf
  https://arxiv.org/abs/1703.09934}

\bibitem{Christiansson2007}
G.~A.~V. Christiansson, {Hard Master, Soft Slave Haptic Teleoperation}, Ph.D.
  thesis, Delft University of Technology (2007).

\bibitem{Aiple2019b}
M.~Aiple,
  \href{http://doi.org/10.4121/uuid:d3309b3b-dce7-4f1a-bfd7-0113646152c3}{{Instructional
  video to the self-adapting variable impedance actuator experiment}} (2019).
\newblock \href
  {https://doi.org/doi:10.4121/uuid:d3309b3b-dce7-4f1a-bfd7-0113646152c3}
  {\path{doi:doi:10.4121/uuid:d3309b3b-dce7-4f1a-bfd7-0113646152c3}}.
\newline\urlprefix\url{http://doi.org/10.4121/uuid:d3309b3b-dce7-4f1a-bfd7-0113646152c3}

\bibitem{Hashtrudi-zaad2001}
K.~Hashtrudi-Zaad, S.~E. Salcudean,
  \href{http://journals.sagepub.com/doi/abs/10.1177/02783640122067471
  http://journals.sagepub.com/doi/abs/10.1177/02783640122067471}{{Analysis of
  Control Architectures for Teleoperation Systems with Impedance/Admittance
  Master and Slave Manipulators}}, The International Journal of Robotics
  Research 20~(6) (2001) 419--445 (6 2001).
\newblock \href {https://doi.org/10.1177/02783640122067471}
  {\path{doi:10.1177/02783640122067471}}.
\newline\urlprefix\url{http://journals.sagepub.com/doi/abs/10.1177/02783640122067471
  http://journals.sagepub.com/doi/abs/10.1177/02783640122067471}

\bibitem{Seborg2016}
D.~E. Seborg, T.~F. Edgar, D.~A. Mellichamp, F.~J. Doyle~III, {Process Dynamics
  and Control}, 4th Edition, Wiley, 2016 (2016).

\bibitem{Wilcoxon1945}
F.~Wilcoxon,
  \href{https://www.jstor.org/stable/10.2307/3001968?origin=crossref}{{Individual
  Comparisons by Ranking Methods}}, Biometrics Bulletin 1~(6) (1945) 80 (12
  1945).
\newblock \href {https://doi.org/10.2307/3001968} {\path{doi:10.2307/3001968}}.
\newline\urlprefix\url{https://www.jstor.org/stable/10.2307/3001968?origin=crossref}

\bibitem{Aiple2020b}
M.~Aiple,
  \href{http://doi.org/10.4121/uuid:6b745205-e38f-4190-9460-bea1b53d50b0}{{Human
  performing precision and dynamic tasks through a 1-DOF teleoperator with
  self-adapting variable impedance actuator}} (2020).
\newblock \href
  {https://doi.org/doi:10.4121/uuid:6b745205-e38f-4190-9460-bea1b53d50b0}
  {\path{doi:doi:10.4121/uuid:6b745205-e38f-4190-9460-bea1b53d50b0}}.
\newline\urlprefix\url{http://doi.org/10.4121/uuid:6b745205-e38f-4190-9460-bea1b53d50b0}

\bibitem{Aiple2017a}
M.~Aiple,
  \href{https://data.4tu.nl/repository/uuid:32cb858b-c37f-4e25-be0d-b0ea6c5c2b86}{{Instructional
  video to the teleoperated flexible hammering experiment}} (2017).
\newblock \href
  {https://doi.org/10.4121/uuid:32cb858b-c37f-4e25-be0d-b0ea6c5c2b86}
  {\path{doi:10.4121/uuid:32cb858b-c37f-4e25-be0d-b0ea6c5c2b86}}.
\newline\urlprefix\url{https://data.4tu.nl/repository/uuid:32cb858b-c37f-4e25-be0d-b0ea6c5c2b86}

\bibitem{Pratt1997}
G.~A. Pratt, M.~M. Williamson, P.~Dillworth, J.~Pratt, A.~Wright,
  \href{http://link.springer.com/10.1007/BFb0035216}{{Stiffness isn't
  everything}}, in: O.~Khatib, J.~K. Salisbury (Eds.), Experimental Robotics
  IV, Springer-Verlag, London, 1997, pp. 253--262 (1997).
\newblock \href {https://doi.org/10.1007/BFb0035216}
  {\path{doi:10.1007/BFb0035216}}.
\newline\urlprefix\url{http://link.springer.com/10.1007/BFb0035216}

\end{thebibliography}
\end{document}